\documentclass[10pt,twocolumn,letterpaper]{article}

\usepackage[pagenumbers]{cvpr} % To force page numbers, e.g. for an arXiv version
\usepackage[utf8]{inputenc}
\usepackage[ruled,longend]{algorithm2e}
\usepackage{textgreek}
\usepackage{amssymb}
\usepackage{tcolorbox}
\usepackage{algpseudocode}
\usepackage{multirow}
\usepackage{appendix}
\usepackage{pythonhighlight}
\usepackage[at]{easylist}

\definecolor{Gray}{gray}{0.9}
\definecolor{cvprblue}{rgb}{0.21,0.49,0.74}
\usepackage[pagebackref,breaklinks,colorlinks,allcolors=cvprblue]{hyperref}

\def\paperID{7957} % *** Enter the Paper ID here
\def\confName{CVPR}
\def\confYear{2025}

\title{SubZero: Composing \underline{Sub}ject, Style, and Action via \underline{Zero}-Shot Personalization}

\author{
Shubhankar Borse \thanks{Corresponding Author} \quad
Kartikeya Bhardwaj \thanks{These authors contributed equally to this work.}
\quad
Mohammad Reza Karimi Dastjerdi \footnotemark[2]
\quad
Hyojin Park \footnotemark[2]
\and
Shreya Kadambi
\quad
Shobitha Shivakumar
\quad
Prathamesh Mandke
\quad
Ankita Nayak
\quad
Harris Teague
\and
Munawar Hayat \footnotemark[1]
\quad
Fatih Porikli\\
{Qualcomm AI Research \thanks{Qualcomm AI Research, an initiative of Qualcomm Technologies, Inc.} }\\
{\tt\small \{sborse, hayat\}@qti.qualcomm.com}\\
}
\begin{document}
\maketitle
\begin{abstract}

%This work proposes a training-free approach to diffusion based personalization. 
%We strive for optimal content-style composition and develop a highly synergized set of constraints to fuse content and style information without any leakage.
%We leverage from domain-expert models specialized in style-description and subject classification to faithfully encode the desired style characteristics while preserving subjects identity during the generation process. 

Diffusion models are increasingly popular for generative tasks, including personalized composition of subjects and styles. While diffusion models can generate user-specified subjects performing text-guided actions in custom styles, they require fine-tuning and are not feasible for personalization on mobile devices. Hence, tuning-free personalization methods such as IP-Adapters have progressively gained traction. However, for the composition of subjects and styles, these works are less flexible due to their reliance on ControlNet, or show content and style leakage artifacts. To tackle these, we present SubZero, a novel framework to generate any subject in any style, performing any action without the need for fine-tuning. We propose a novel set of constraints to enhance subject and style similarity, while reducing leakage. Additionally, we propose an orthogonalized temporal aggregation scheme in the cross-attention blocks of denoising model, effectively conditioning on a text prompt along with single subject and style images. We also propose a novel method to train customized content and style projectors to reduce content and style leakage. Through extensive experiments, we show that our proposed approach, while suitable for running on-edge, shows significant improvements over state-of-the-art works performing subject, style and action composition. 

\vspace{- 1.2 em}

% Diffusion models have become increasingly popular for performing generative tasks, including personalized composition of subjects and styles. PEFT-based methods are utilized for generating custom objects performing text-driven actions in custom styles. However, these methods require fine-tuning and are not feasible to use for personalization on mobile devices.
% Hence, tuning-free personalization methods such as IP-Adapters have progressively gained traction. However, for the composition of subjects and styles, these works are either limited to less flexibility in text generation process due to their reliance on controlnet, or display content and style leakage artifacts, observed in DDIM inversion. Hence, we present SubZero, a novel framework which given single reference images, generates any subject in any style, performing any action without the need for finetuning. We propose novel latent objective functions to enhance subject and style similarity, while reducing leakage. Additionally, we propose an orthogonalized temporal aggregation scheme in the cross-attention blocks of denoising networks, effectively conditioning on a text prompt along with single subject and style images. Furthermore, we propose a novel method to train customized content and style projectors in a way which reduces content and style leakage. Our extensive experiments show a tremendous improvement over previous works performing subject, style and action composition. Finally, we show that the SubZero algorithm is feasible to run on a mobile device without the need for online compute. 

\end{abstract}    
\section{Introduction}
\label{sec:intro}

\begin{figure}[t]
    \centering
    \includegraphics[width=1.0\linewidth]{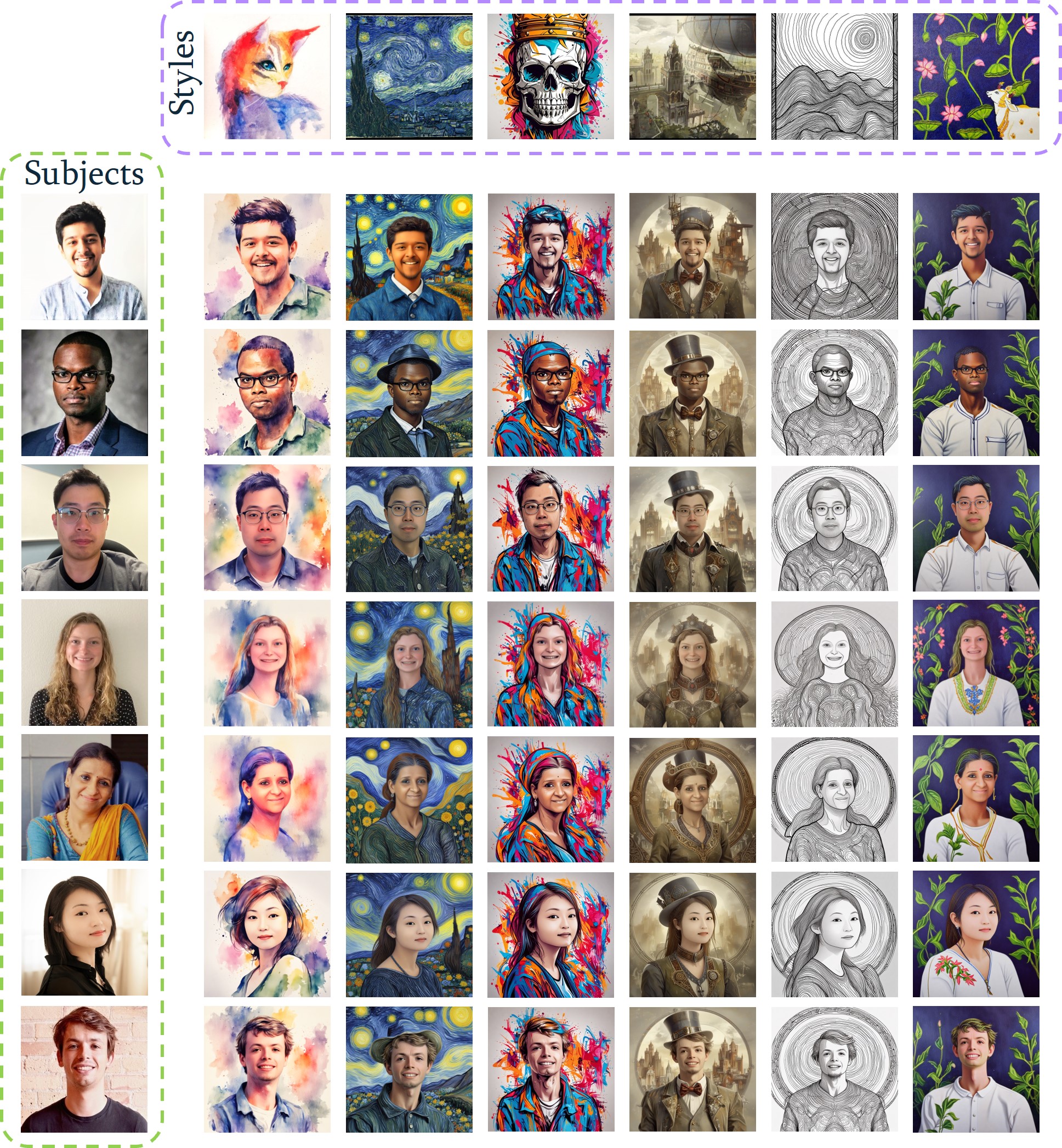}
    \vspace{- 2 em}
    \caption{
    % \textbf{Stylized Faces with SubZero}. Stylized Faces with SubZero.
Various stylized face images generated using our proposed SubZero method applied to pre-trained text-to-image diffusion models without any tuning. 
SubZero produces high-quality, diverse stylized images while maintaining facial features.
    }%, and suffer from mode collapse. However, our proposed FouRA produces more diverse images.}
    \label{fig:faces_stylized}
   \vspace{- 1.2 em}
\end{figure}

Large Text-to-Image (T2I) generative models based on diffusion have gained traction~\cite{ddpm, podellsdxl, rombach2022high}, surpassing other existing methods \cite{goodfellow2014generative}. While these models can generate high-fidelity and diverse images\cite{esser2024scaling}, gaining control over synthesized images by ensuring consistent subjects or styles remains a significant challenge~\cite{blora, shah2025ziplora}.
%Stable Diffusion has revolutionized computer vision by bringing the ability to generate novel images to the masses. One of the key developments in this field are Generative AI methods that allow users to personalize images either to contain a specific subject (e.g., a specific person or common objects/pets used in DreamBooth~\cite{ruiz2023dreambooth}) or by editing generated images to have certain styles~\cite{hertz2024style, ye2023ip-adapter}. For example, successful style transfer is commonly accomplished using the popular Low Rank Adaptation (LoRA) technique~\cite{lora} and its variants~\cite{shira, foura, blora}. Recently, \textit{subject-style composition} has emerged which integrates both subject and styles into text-to-image diffusion models~\cite{shah2025ziplora,gu2024mix,blora,wang2024instantstyle}. These methods either (\textit{i})~rely on training subject- or style-specific LoRAs and then blending them together (e.g., ZipLoRA~\cite{shah2025ziplora}, Mix-of-Show~\cite{gu2024mix}, B-LoRA~\cite{blora}), or (\textit{ii})~exploit expensive models architectures (e.g., InstantStyle-Plus~\cite{wang2024instantstyle}, ControlNet~\cite{controlnet}), or DDIM inversion (e.g., StyleAligned~\cite{hertz2024style}) to directly work with a single reference image, or (\textit{iii})~leak irrelevant content from reference images into the generated images (e.g., IP-Adapter~\cite{ye2023ip-adapter}). 

To address this issue, recent studies have proposed fine-tuning diffusion models using reference images~\cite{blora, shah2025ziplora, gu2024mix, foura, shira}. 
They utilize LoRA~\cite{lora} for efficient training while preserving original models capability. While this approach has demonstrated a remarkable ability to control the style or content of generative model, it lacks generalization and requires availability of multiple training samples incurring additional memory and time for adaptation. 
Moreover, these methods require fine-tuning a dedicated adapter each time we need to support new styles or subject images, which is a significant drawback for resource-constrained on-device applications. This key limitation has led to an emergence of training-free methods that can generalize to any reference subject or style images.

Recent training-free methods for \textit{subject-style composition} rely on DDIM inversion-based approaches~\cite{wang2024instantstyle_plus, hertz2024style}, ControlNet-based methods~\cite{controlnet, wang2024instantstyle}, and shared attention techniques~\cite{rout2024rb, hertz2024style}. 
These methods eliminate the need for fine-tuning a different adapter for each subject/style but struggle to properly disentangle content and style information or to preserve subject fidelity. For instance, the DDIM inversion-based methods adapt the noise from the subject image by injecting style information, which can lead to subject leakage from the style image. 
ControlNet-based methods offer good personalization but lack flexibility. Both DDIM inversion and ControlNet based methods perform poorly on generating a diverse range of images. Hence, they also fail when action prompts are added. Moreover, both the techniques are computationally expensive. Other methods such as IP-Adapter~\cite{ye2023ip-adapter} are efficient. However, all the above methods result in \textit{irrelevant subject leakage} (e.g., background from reference subject images leaking into generated images). To tackle subject leakage, RB-modulation~\cite{rout2024rb} elegantly proposed the stochastic optimal control scheme which directly optimizes the diffusion latent. However, our experiments show that RB-modulation fails to effectively align the content with style in the loss and hence results in irrelevant subject leakage. This has also been recently observed by the community~\cite{RB-Issues}.

To enable effective and privacy-preserving subject-style composition on-edge devices, we aim to create a \textit{robust yet efficient} subject, style and action composition method that can (\textit{i})~clearly \textit{disentangle} the subject and style, (\textit{ii})~generate a wide range of images controlled by the text prompt, (\textit{iii})~work with just a \textit{single reference} subject and/or style image instead of training a new adapter for each scenario, and (\textit{iv})~reduce irrelevant subject leakage (e.g., background from subject reference image) into the generated image. 
We propose SubZero, a robust zero-shot solution to subject, style and action composition. At the core of our approach is a novel latent modulation objective formulation, orthogonal and temporally-adaptive blending of subject and style information inside the cross-attention modules, generalized adapters trained to specifically disentangle subjects and styles while limiting irrelevant leakage. With these new ideas, we show high quality subject, style and action composition and face personalization applications (e.g., see Fig.~\ref{fig:faces_stylized}) that are particularly suited for efficient execution due to their low compute costs.

Overall, we make the following \textbf{key contributions}:
\begin{enumerate}
    \item We propose SubZero, a robust \underline{Sub}ject-Style Composition framework with \underline{Zero} training for new concepts.
    \item We propose the disentangled stochastic optimal controller containing novel latent modulation objectives that effectively align subject and style during inference. 
    \item We propose a temporally-adaptive and orthogonal aggregation method to effectively combine attention features originating from subject, style and text conditioning.
    \item We train custom subject and style adapters with novel training techniques and losses, and demonstrate how these new adapters significantly limit irrelevant content leakage compared to the prior art.
    \item Our extensive experiments clearly set a new state-of-the-art on subject-style composition (e.g., for objects such as items or pets, as well as face personalization) as well as subject-style-action composition.
    %\item Finally, we discuss how SubZero can be applied in a completely gradient-free manner in the future using Zero-Order training and provide a proof-of-concept.
\end{enumerate}

%This paper is organized as follows. Section~\ref{sec:related} discusses the related work while Section~\ref{sec:method} proposes our technique. Then, we conduct extensive experiments in Section~\ref{sec:exp} and conclude the paper in Section~\ref{sec:conc}.

% Training free
% DDIM inversion : contents leakage from style (RB mention in paper)
% Intermediate control (controlnet) methods provide good personalization but less flexibility cannot 
% Shared-Attention or CLIPspace methods provide good flexibility but less personalization
% RB module/ dreambooth -> content,style together 
% Advantage Disadvantage
% Good style but cannot keep the content information
% ur solution
% similairt (Dino, other obj, face sim for human)

\section{Related Work}
\label{sec:related}

Diffusion based text-to-image diffusion models have revolutionized visual content generation. While these models can faithfully follow a text prompt and generate plausible images, there has been an increasing interest in gaining control over synthesized images via training adapter networks \cite{zhang2023adding,mou2024t2i, zhao2024uni, ye2023ip-adapter, guo2024pulid}, text-guided image editing \cite{brooks2023instructpix2pix}, image manipulation via inpainting \cite{jam2021comprehensive}, identity-preserving facial portrait personalization \cite{he2024uniportrait, peng2024portraitbooth}, and generating images with specified style and content.

\begin{figure*}[t]
    \centering
    \includegraphics[width=0.75\linewidth]{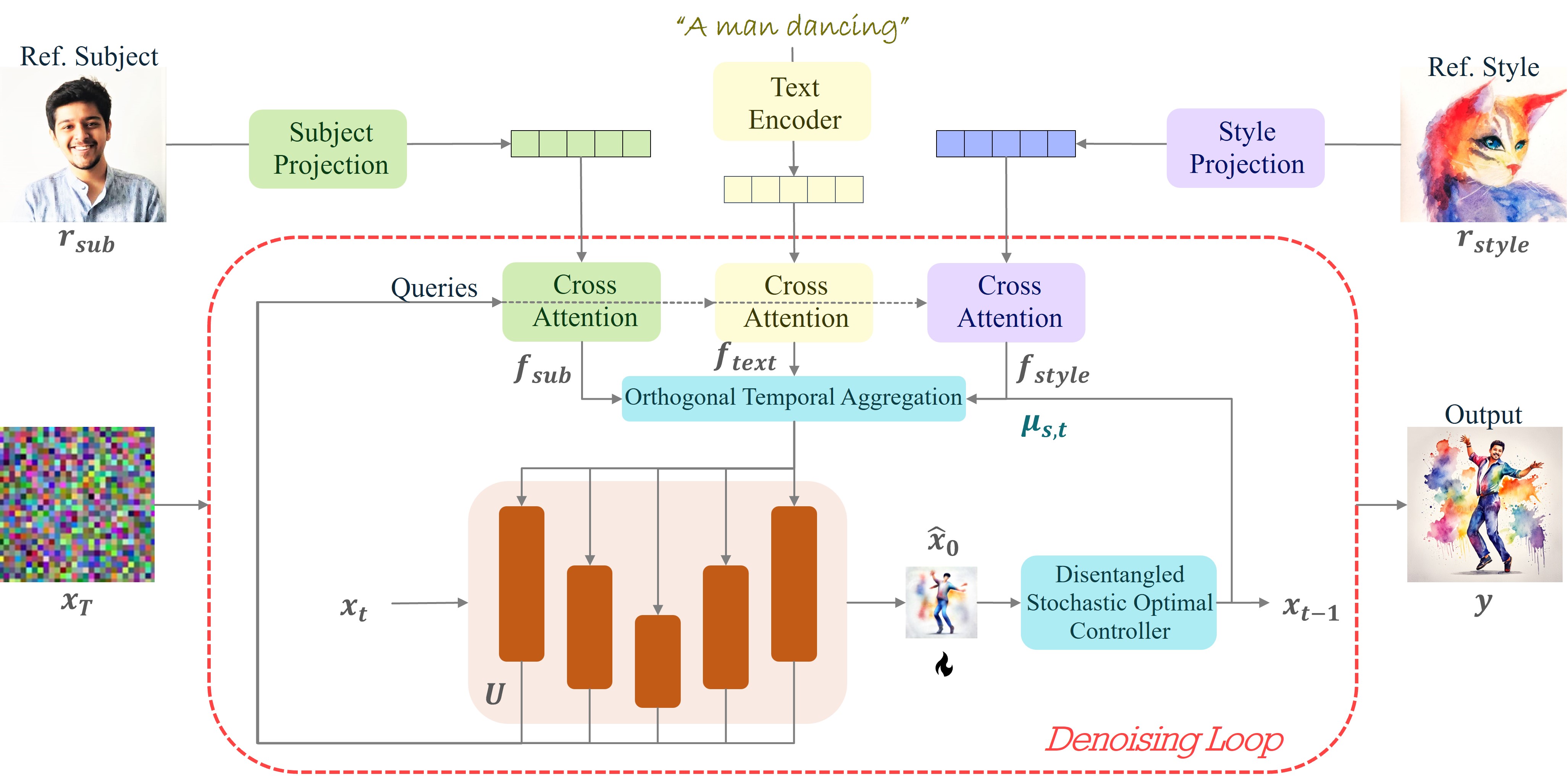}
    %\vspace{- 1.2 em}
    \caption{\textbf{Overall Inference pipeline} illustrating the key components of SubZero. Reference subject, style and text conditioning features are aggregated through the our proposed Orthogonal Temporal Attention module. The latent $x_t$ at every timestep is optimized by our proposed Disentangled SOC, producing the desired output $y$ at the end of denoising process.}
    \label{fig:inference_pipe}
    \vspace{- 0.5 em}
\end{figure*}

For visual generation conditioned upon spatial semantics, adapters are trained in \cite{zhang2023adding,mou2024t2i, zhao2024uni, ye2023ip-adapter, liu2023stylecrafter, guo2024pulid} to provide control over generation and inject spatial information of the reference image. ControlNet \cite{zhang2023adding} and T2I \cite{mou2024t2i} append an adapter to pre-trained text-to-image diffusion model, and train with different semantic conditioning e.g., canny edge, depth-map, and human pose. Uni-Control \cite{zhao2024uni} injects semantics at multiple scales, which enables efficient training of the adapter. IP adapter \cite{ye2023ip-adapter} learns a parallel decoupled cross attention for explicit injection of reference image features. Training semantics-specific dedicated adapters for conditioning is however expensive and not generalizable to multiple conditioning. 

Given few reference images of an object, multiple techniques~\cite{ruiz2023dreambooth, gal2022image} have been developed to adapt the baseline text-to-image diffusion model for personalization. 
Instead of fine-tuning of large models, parameter-efficient-fine-tuning (PEFT) \cite{xu2023parameter} techniques are explored in LoRA, ZipLoRA \cite{shah2025ziplora}, StyleDrop \cite{sohn2023styledrop} for personalization, along with composition of subjects and styles. 
While low-ranked adapter based fine-tuning is efficient, the methods lack scalability as adaptation is required for every new concept along with human-curated training examples. Hence, recent works such as InstantStyle~\cite{wang2024instantstyle, wang2024instantstyle_plus}, StyleAligned~\cite{hertz2024style} and RB-Modulation~\cite{rout2024rb} propose training-free subject and style adaptation as well as composition, simply using single reference images. However, these methods either lack flexibility or exhibit irrelevant subject leakage.

Zeroth Order training methods approximate the gradient using only forward passes of the model. Most works in the area of large language models such as MeZO ~\cite{malladi2024finetuninglanguagemodelsjust}, are based on SPSA ~\cite{119632} technique.
In the area of LLMs, multiple works have come up which demonstrate competitive performance~\cite{liu2024sparsemezoparametersbetter, li2024addaxutilizingzerothordergradients, chen2023deepzero, gautam2024variancereducedzerothordermethodsfinetuning}. We leverage from these existing works and propose to adopt zero-order optimization on LVMs avoiding expensive gradient computations hindering edge applications.
%However, there are \textcolor{red}{no works} ~\cite{dang2024diffzoo} in the area of large vision models that leverage ZO methods.%, that we are aware of.

\vspace{- 0.7 em}
\section{Proposed Approach}
\label{sec:method}

%In this Section, we provide a detailed description of our approach. We start with the method preliminaries in \ref{subsec:background}, followed by our proposed Disentangled Stochastic Optimal Controller and orthogonal Temporal Aggregation schemes for subject and style composition. While SubZero works out-of-the-box on existing adapters, we provide further insight into training targeted projectors for object and style composition. 

In this Section, we provide a detailed description of our approach. We briefly summarize preliminaries in Sec.\ref{subsec:background}. In Sec.\ref{subsec:dsoc}, we elaborate on the Disentangled Stochastic Optimal Controller to reduce subject and style leakage while preserving identity. To further facilitate effective information composition, we propose orthogonal Temporal Aggregation schemes in Sec.\ref{subsec:ota}. While SubZero works out-of-the-box on existing adapters, we provide additional insight into training targeted projectors for object and style composition in Sec. \ref{subsec:train}. Finally, we propose an extension of our work to Zero-Order Stochastic Optimal Control in Sec.~\ref{subsec:zo}.

%In the following, we provide a detailed description of our approach. We start with the method overview, followed by our proposed latent objective functions, with further discussions on our adaptive weighted attention strategy.

\subsection{Preliminaries}
\label{subsec:background}
\vspace{- 0.2 em}

\textbf{Text-to-Image Generation:} Diffusion-based models such as~\cite{rombach2022high, podellsdxl, perniaswurstchen} are widely adopted for text-to-image generation. As they usually require 20-30 inference steps, recent works such as~\cite{lin2024sdxl} have also been adopted to speed up their latent denoising process. Our approach is developed on two efficient foundational models: SDXL-Lightning~\cite{lin2024sdxl} (4-step) and W\"{u}rstchen~\cite{perniaswurstchen}. The goal is to model a denoising operation given a forward noising process:
\begin{equation}
\label{eq:fwdproc}
    x_t = \sqrt{\alpha_t}x_0 + \sqrt{1-\alpha_t}\epsilon, \quad  \epsilon \sim N(0, 1)
\end{equation}

Here, $x_t$ represents the state at time $t \in [0, \infty) $, given the original input $x_0$, and $\alpha_t$ is computed by a scheduler.

Current methods~\cite{rombach2022high, podellsdxl, perniaswurstchen} are developed with the objective of reversing the equation~\ref{eq:fwdproc}. They consist of an Encoder-Decoder model $\mathbf{V_e}, \mathbf{V_d}$ which transforms images to and from the latent representation $x_t$, and denoising model $\mathbf{U}$ which progressively de-noises input latents to estimate the noise at every timestep. For SDXL, we denote the Unet as $\mathbf{U}$, and VAE decoder as $\mathbf{V_d}$. For W\"{u}rstchen, we denote the StageC denoiser and the StageA VAE as $\mathbf{U}$ and $\mathbf{V_d}$ respectively. To produce a text-conditioning for the denoising model, the text prompt $\mathbf{p}$ is tokenized and encoded via a text encoder $\bf{\phi_p}$ (i.e. clip \cite{clip}). The output embeddings are fed to $\mathbf{U}$ as keys and values in stage-wise cross-attention modules. The queries to each cross-attention module are the intermediate latent features from $\mathbf{U}$.

\noindent \textbf{Stochastic Optimal Control:} RB-Modulation~\cite{rout2024rb} recently developed latent optimization with stochastic optimal control to effectively adapt intermediate latents produced by $\mathbf{U}$ to inject a reference style $r_{sty}$. For accurate measurement of style, they used the Contrastive Style Descriptor (CSD) network~\cite{somepalli2024measuring} $\mathbf{\psi}$. To perform stochastic optimal control, the intermediate latent $x_t$ at timestep $t$ is used to predict de-noised latent $\hat{x}_0$ as follows:
\begin{equation}
\label{eq:pred_x0}
    \hat{x}_0 = \frac{x_t}{\alpha_t} + \frac{(1-\sqrt{\bar{\alpha_t}})}{\sqrt{\bar{\alpha_t}}}\mathbf{U}(x_t, t, \mathbf{p});
\end{equation}

Keeping only $\hat{x}_0$ as tunable, the denoised image is predicted as $\hat{y} = \mathbf{V_d}(\hat{x_0})$. A style objective $\mathcal{L} = \|\mathbf{\psi} (\hat{y}) - \mathbf{\psi} (r_{sty})\|^{2}_{2}$ is then computed as the terminal cost.
% $\ell_\mathrm{style} = \|\mathbf{\psi} (\hat{y}) - \mathbf{\psi} (y)\|^{2}_{2}$
Finally, the Adam optimizer is used to update $\hat{x_0}$ to reduce the style objective for $M$ iterations. The updated $\hat{x_0}$ is now used to compute denoised latent for the previous time-step $x_{t-1}$. 

\noindent \textbf{Reference Image Conditioning:}
To condition the denoising model using reference subject image $r_{sub}$ and style image $r_{sty}$, there have been various lines of work. For example, training additional customized key and value projections in the cross-attention blocks of $\mathbf{U}$ for reference images of concepts, such as IP-Adapter~\cite{ye2023ip-adapter} and PulID~\cite{guo2024pulid}. Another line of work, such as the Attention Feature Aggregation (AFA) proposed by RB-Modulation, pass the reference image through the clip-image encoder $\bf{\phi_i}$ to encode reference images, and use the key/value projections already available in the base model for conditioning. This method is however, native only to the W\"{u}rstchen model, as it contains already learnt clip text and image projectors. Hence for fair comparison with all baselines, we use IP-Adapter-based projections to encode reference conditions in SDXL experiments, and AFA-based conditioning in W\"{u}rstchen \cite{perniaswurstchen}.

For the methods discussed above, queries from $\mathbf{U}$ are attended separately by key-value projections from all modalities (text, style, subject) or an aggregation of key-value projections in these modalities. In our work, we denote the updated features as $f_{text}$, $f_{style}$ and $f_{sub}$. After feature aggregation, the updated features after aggregating cross-attention outputs from all modalities is denoted as $f_{agg}$. 

\subsection{Disentangled Stochastic Optimal Controller}
\label{subsec:dsoc}
% 
% \textbf{Preliminaries}
% For T2I generation, let $s(x_t; t,\theta)$ be the neural network from which we sample the score function of the joint predictive distribution $\nabla p(X_t;y)$. For personalization in diffusion, we follow the optimal control method in RB modulation. Given a reference image $X^f_0$, and sampled latent from the diffusion process $X^u_t$, terminal cost is defined as the l2 term 
% \begin{math}
% \centering
% \mathcal{L}(x_0) = ||\psi(z_0) - \psi(x_0)||^{2}_{2}
% \end{math}, 
% where the style features are computed by the Contrastive Style Descriptor (CSD) model $\Psi(.)$~\cite{somepalli2024measuring}.

RB Modulation showed that direct feature injection can cause subject leakage from style reference images.  However, our studies show that the stochastic optimal controller and AFA modules are not able to alleviate the subject leakage problem. This has also been observed by the community~\cite{RB-Issues}. Additionally, the approach is not able to preserve necessary characteristics of faces for face personalization (see Fig.~\ref{fig:wurschten_comp}). 
Hence, we propose the Disentangled Stochastic Optimal Controller to alleviate subject and style leakage, while preserving key features of the subjects along with styles. Algorithm~\ref{alg:inference} provides pseudo-code for the proposed Disentangled Stochastic Optimal Controller.

\noindent \textbf{Subject and Style Descriptors}: As discussed in Sec.~\ref{subsec:background}, RB-Modulation optimizes latents for style descriptor $\psi$. Their terminal cost however does not take into account the personalized features of the subject image. Hence, we propose an additional term for personalization of the reference image, computed by a subject descriptor $\rho$. For face stylization experiments, we replace $\rho$ with a facial descriptor $\delta$. Throughout this paper, we use style descriptors $\psi$ from the CSD network~\cite{somepalli2024measuring}, the subject descriptor network as DINO~\cite{Caron_2021_ICCV}, and the facial descriptor as $\delta$ as the facial embedding extractor trained by~\cite{ye2023ip-adapter}, using  Arc-Face~\cite{deng2019arcface}. %We provide further details of all the descriptors in  \textcolor{red}{ Section~\ref{}}.

\begin{figure}[h]
    \centering
    \includegraphics[width=0.85\linewidth]{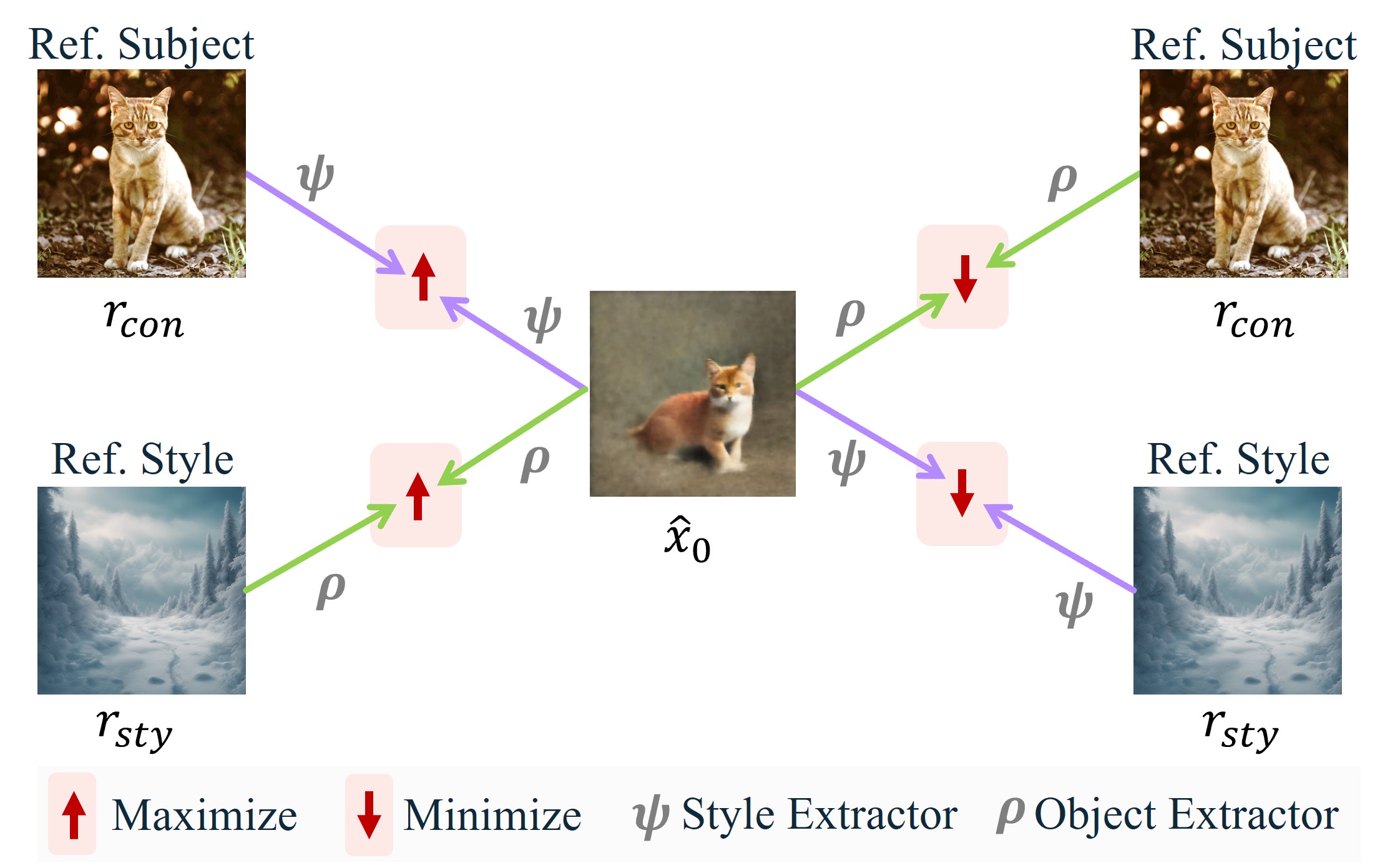}
    %\vspace{- 1.2 em}
    \caption{\textbf{Disentangled Stochastic Optimal Controller}.}%, and suffer from mode collapse. However, our proposed FouRA produces more diverse images.}
    \label{fig:latent_objectives}
    \vspace{- 1 em}
\end{figure}

%We add the maximizing the distance between the style descriptors of the reference concept image $r_con$. In addition, to minimize the leakage of style from the reference content image we apply 
We also propose negative criteria aiming to reduce content and style leakage between networks. This is achieved by maximizing descriptors from $\rho$ for $r_{sty}$ and maximizing descriptors from $\psi$ for $r_{con}$. The terminal cost is hence a combination of four objectives, see Fig.~\ref{fig:latent_objectives}.

\noindent \textbf{Terminal Cost}: We define the terminal cost $\mathcal{L}$ as, 
% \vspace{-0.5 em}
\begin{align}
\begin{split}
\label{eq:dsoc}
    \mathcal{L} = \underbrace{\|\mathbf{\rho} (\hat{y}) - \mathbf{\rho} (r_{sty})\|^{2}_{2}}_{\text{subject descriptor constraint } \mathcal{L}_{c}} 
     + \underbrace{\|\mathbf{\psi} (\hat{y}) - \mathbf{\psi} (r_{sub})\|^{2}_{2}}_{\text{style descriptor constraint } \mathcal{L}_{s}} \\ 
     -\gamma_{nc}\underbrace{\|\mathbf{\psi} (\hat{y}) - \mathbf{\psi} (r_{sty})\|^{2}_{2}}_{\text{subject leakage constraint } \mathcal{L}_{nc}} -\gamma_{ns} \hfill\underbrace{\|\mathbf{\rho} (\hat{y}) - \mathbf{\rho} (r_{sub})\|^{2}_{2}}_{\text{style leakage constraint } \mathcal{L}_{ns}}
\end{split}
 \vspace{-0.8 em}
\end{align}

where $\hat{y}$ is the estimated denoised image $\mathbf{V_d}(\hat{x}_0)$, $\mathbf{V_d}$ is a TinyVAE decoder~\cite{TinyVAE}, $\gamma_{ns}$ and $\gamma_{nc}$ are weighting terms for style and content leakage and are used as hyperparameters, their values are provided in the Appendix.

% \begin{align}
% \begin{split}
%     \mathcal{L}(x_0) = \underbrace{min_{u \in \mathcal{U}} ||\Gamma(z_0) - \Gamma(X^u_0)||^{2}_{2}}_{\text{facial descriptor constraint}}  \\
%      + \hfill \underbrace{min_{u \in \mathcal{U}} ||\Phi(z_0) - \Phi(X^u_0)||^{2}_{2}}_{\text{concept descriptor constraint}} 
% \end{split}
% \end{align}
%Where, $\Phi(.)$ provides the concept embedding and the   $\Gamma(.)$ provides the facial descriptor embedding. We use CLIP Image encoder to extract visual features further project it using a QFormer model initialized using IP-Adapter weights to extract facial features

%To maintain high fidelity of facial content and the concept, We extend the controller cost $\mathcal{L}(x_0)$, to add the concept criterion as 

% \begin{math}
% \mathcal{L}(x_0) := ||\psi(z_0) - \psi(x_0)||^{2}_{2} + \lambda ||x_0 - \mathrm{E}[X^{u}_{0}|X^{u}_{t}]||^{2}_{2} + ||\partial(z_0) - \partial(x_0)||^{2}_{2}
% \end{math}

% \subsection{Latent Objective Functions}
% RB Modulation, showed that direct feature injection can cause content leakage from style reference images and introduced Attention feature aggregation (AFA).  AFA module decouples style from the content, we show that it cannot retain concept/face alignment. We show that adding a negative terminal cost in addition to the cost introduced by the face and style descriptors can alleviate leakage problem further. 
% We add the maximizing the distance between the style descriptors of the reference concept image $r_con$. In addition, to minimize the leakage of style from the reference content image we apply 

%\subsection{Adaptive Weighted Attention}

\subsection{Orthogonal Temporal Attention Aggregation}
\label{subsec:ota}
As discussed in Section~\ref{subsec:background}, within our denoising model $\mathbf{U}$, we obtain the updated features $f_{text}$, $f_{style}$ and $f_{sub}$ from three sources of conditioning after cross attention. Previous works~\cite{rout2024rb, ye2023ip-adapter} have proposed a weighted addition of these updated features, to obtained aggregated features $f_{agg}$. However, we observe that this leads to subject leakage in the generated image, as discussed in the Appendix.

% \noindent \textbf{Orthogonal features:} Subject features need to update local regions of the latent. Hence, SubZero uses an orthogonal projection of the subject query, $\hat{f}_{sub}$, over the original text features. This preserves key aspects of the text prompt, such as any actions described for the subject and generates robust images depending on text conditioning. On the other hand, style is a global construct. Hence we directly add the style query without orthogonalization, allowing the style condition to update the image holistically instead of a local area. 

\begin{figure}[ht]
    \centering
    \vspace{- 0.5 em}
    \includegraphics[width=0.7\linewidth]{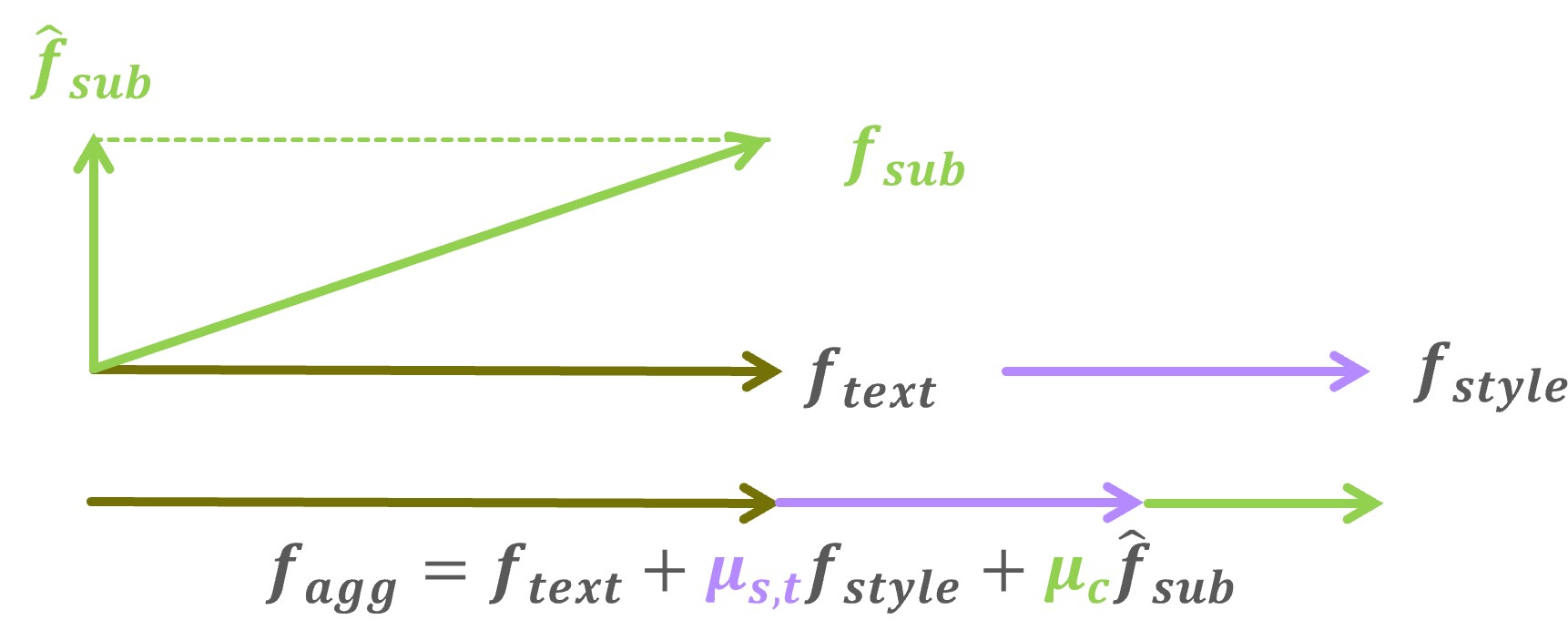}
    \caption{\textbf{Orthogonal Temporal Aggregation}.}
    \label{fig:orthogonal_queries}
    \vspace{- 0.5 em}
\end{figure}

\noindent \textbf{Orthogonal features:} 
The text and style features contribute to the global structure, while the subject features update local regions of the latent space. 
To prevent distortion between various sources of information in the latent space, we apply an orthogonal projection of the subject query, $\hat{f}_{sub}$, onto the original text  to update local regions. 
Meanwhile, the style query is directly added to the text features to update the image holistically, as shown in Fig \ref{fig:orthogonal_queries}. 
This approach preserves key aspects of each component, such as actions described for the subject in the text prompt, and generates robust images based on text and image conditioning.

\noindent \textbf{Temporal Weighting:} To reduce the subject leakage problem, we propose a temporal weighting strategy. To weigh the updated queries, we use a novel temporal-adaptive weighting mechanism. As style is a global construct, it should not decide the shape of objects generated in the image. The shapes should be decided based on text-conditioning features and subject-conditioning features. Hence, at the start of the denoising process, when shapes are being generated, we fix a lower weight for style features $f_{style}$ and a higher scale for subject features $f_{sub}$. As the denoising process progresses, we increase the style scale gradually based on two factors: direct proportionality to the style descriptor constraint $\mathcal{L}_{s}$ and inverse proportionality to the subject leakage constraint $\mathcal{L}_{ns}$, determined in Equation~\ref{eq:dsoc}. At timestep $t$, the temporal style weights are denoted as $\mu_{s,t}$, and subject weights are denoted as $\mu_{c}$. Algorithm~\ref{alg:inference} provides pseudo-code for $\mu_{s,t}$. 

Finally, the Orthogonal Temporal Aggregation (OTA) features are calculated as $f_{agg} = f_{text} + \mu_{s,t}f_{style} + \mu_{c}f_{sub}$.

\begin{algorithm}[ht]
\caption{SubZero: Disentangled Controller and Temporal Aggregation}
\label{alg:inference}
%\begin{algorithmic}[1] 
\textbf{Input}: Reference subject image $r_{sub}$, reference style image $r_{sty}$, style descriptor $\psi$, Subject extractor $\rho$, text prompt $\mathbf{p}$, Denoising Network $\mathbf{U}$, TAE decoder $\mathbf{V_d}$ \\
\textbf{Tunable Parameter}: Step size $\eta$, Optimization steps $M$,  Initial style scale $\mu_{s,0}$, Style tuner $\zeta$ \\
\SetAlgoLined
Initialize $x_T \gets \mathcal{N}(0,I_d)$;

\For{t=T to 1}
{
    \textit{Compute Predicted latent}: \\ 
    $\hat{x}_0 = \frac{x_t}{\alpha_t} + \frac{(1-\sqrt{\bar{\alpha_t}})}{\sqrt{\bar{\alpha_t}}}\mathbf{U}(x_t, t, \mathbf{p})$\;
    \textit{Initialize} $z_0 \to \hat{x}_0$\;
    \For{t=M to 1}
    {
       $\hat{y} = \mathbf{V_d}(\hat{x}_0)$\;
       \textit{Compute disentangled control objective}: \\
        $\mathcal{L} = \mathcal{L}_s + \mathcal{L}_c - \gamma_{nc}\mathcal{L}_{nc} - \gamma_{ns}\mathcal{L}_{ns}$
        $= \|\mathbf{\rho} (\hat{y}) - \mathbf{\rho} (r_{sty})\|^{2}_{2} 
     + \|\mathbf{\psi} (\hat{y}) - \mathbf{\psi} (r_{sub})\|^{2}_{2}
     -\gamma_{nc}\|\mathbf{\rho} (\hat{y}) - \mathbf{\rho} (r_{sty})\|^{2}_{2} -\gamma_{ns}\|\mathbf{\psi} (\hat{y}) - \mathbf{\psi} (r_{sub})\|^{2}_{2}$\;
    \textit{Update optimization variable $z_0$}: \\
    $z_0$ = $z_0 - \eta$ $\nabla_{z_{0}}$ $\mathcal{L}$ $(z_0)$\;
    } 
    $\hat{x}_0 \to z_0$\;
    \textit{Set temporal weighting term:}
    $\mu_{s,t-1} = \mu_{s,t-1} + \zeta\mathcal{L}_s(1-\mathcal{L}_{nc})$\;
    \textit{Compute previous state:}
    $x_{t-1}=DDIM(\hat{x}_0, x_t)$
}
\textbf{Output}: Denoised Image $y = \mathbf{V_d}(x_0)$

%\end{algorithmic}
\end{algorithm}

 % \vspace{- 0.7 em}

% pmandke - WIP

% \begin{tcolorbox}[fonttitle=\bfseries, title=SubZero: Algorithm for latent objective]
% \end{tcolorbox}

% using mezo in latent udpate
% \textbf{ZO for Latent Update:}
% SPSA based ZO methods approximate the gradient by perturbing the weight parameters by a small amount based on some random noise.
% We perform preliminary experiments by leveraging the ZO-Adam scheme described in MeZO ~\cite{malladi2024finetuninglanguagemodelsjust} and extend it to update the latent.
% One of the key challenges with ZO methods is slow convergence primarily resulting from high variance of the noisy gradient approximations.
% In our experiments we observe that while ZO-Adam is not at par with gradient descent, it shows promising performance - achieving a competitive personalization distance given enough training iterations.
% However, the memory savings resulting from ZO-Adam are significant.
% Thus, we suggest the use of ZO techniques for the latent update in scenarios where one can afford to trade training time for a more favorable memory budget.
% Our experiments in ZO are preliminary, and moving forward we intend to explore this area in much more detail.

% [optional] some results: just to show that we achieved decent convergence and that ZO has promise

% @sborse feel free to modify or add more info

\begin{figure*}[t]
    \centering
    \includegraphics[width=0.85\linewidth]{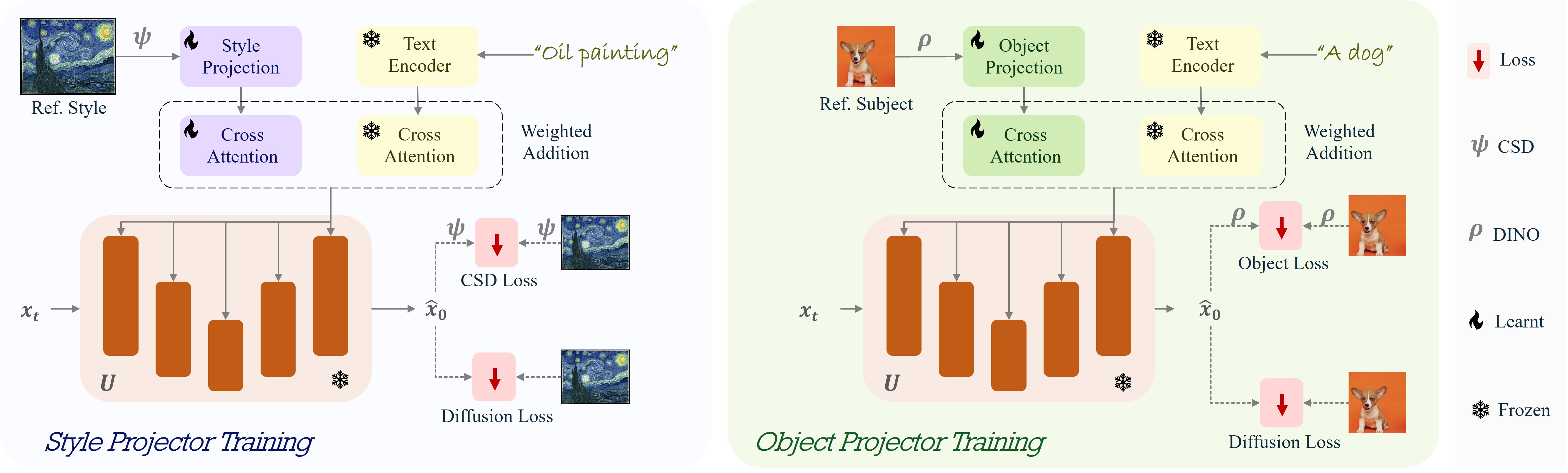}
    %\vspace{- 1.2 em}
    \caption{\textbf{Training Pipeline for StyleZero and ObjectZero projectors}. To train disentangled projectors, we use a weighted combination of the denoising diffusion loss along with a targeted loss to help extract only relevant information from styles and objects.}
    \label{fig:training_pipe}
    \vspace{- 0.5 em}
\end{figure*}

\subsection{Targeted Style and Object Projectors}
\label{subsec:train}
%For content and style, we train two separate adapters that project the DINO features of images into the same space as the text features in the pretrained diffusion model. These adapters are followed by a layer norm and two other projection layers for K and V. The style adapter is trained on the ContraStyles dataset~\cite{somepalli2024measuring}. The content adapter is trained on the MS COCO dataset~\cite{Lin2014MicrosoftCC}. 

While our proposed SubZero algorithm works out-of-the-box on existing IP-Adapters~\cite{ye2023ip-adapter, guo2024pulid, wang2024instantstyle_plus}, we further propose a method to train new style and object projectors. Here, the aim is to disentangle and extract only the relevant information from subjects and styles because IP-Adapters are also known to cause subject leakage. To this end, we utilize the subject and style descriptor models ($\rho$ and $\psi$) to train targeted projectors for objects and styles. 

To train our proposed projectors, we set them as tunable and attach them to every cross-attention block in the denoising model $\mathbf{U}$, which is kept frozen. During each training iteration, we randomly sample the timestep $t$ and compute the noisy latent $x_t$ using the scheduler. We compute the diffusion loss $\ell_\mathrm{denoising}$ on the predicted noise during training.

\noindent\textbf{StyleZero:} We illustrate the training setup for our style projector (StyleZero) in Fig.~\ref{fig:training_pipe}. We use images $y$ from the recent ContraStyles dataset~\cite{somepalli2024measuring} as ground-truth. We first employ the style descriptor (CSD) $\psi$ to extract style embeddings of the reference style image. Next, we pass these descriptors through a Style Projection Network, before passing through key-value projections. These are fed to a cross-attention module, with query projections directly from intermediate features of $\mathbf{U}$. Given noisy image at timestep $t$, we first predict $\hat{x}_0$ using Equation~\ref{eq:pred_x0}. Next, we pass it to the VAE decoder to obtain de-noised prediction $\hat{y} = \mathbf{V_d}(\hat{x_0})$. Similar to the stochastic objective $\mathcal{L}_{s}$, we compute the style loss $\ell_\mathrm{style} = \|\mathbf{\psi} (\hat{y}) - \mathbf{\psi} (y)\|^{2}_{2}$. Hence, the final loss for StyleZero is $\ell_\mathrm{final} = \ell_\mathrm{denoising} + \gamma\ell_\mathrm{style} \ $.

% \begin{equation}
% \ell_\mathrm{final} = \ell_\mathrm{denoising} + \gamma\ell_\mathrm{style} \ .
% \label{eqn:final_style_loss}
% \end{equation}

\noindent\textbf{ObjectZero:} We illustrate the training setup for our object projector (ObjectZero) in Fig.~\ref{fig:training_pipe}. We use images $y$ from MSCOCO~\cite{Lin2014MicrosoftCC} as ground-truth. Similar to StyleZero, we first employ an object descriptor $\rho$ (DINO encoder) to project object embeddings. Similar to the stochastic objective $\mathcal{L}_{c}$, we compute the object loss $\ell_\mathrm{object} = \|\mathbf{\rho} (\hat{y}) - \mathbf{\rho} (y)\|^{2}_{2}$. Hence, the final loss function for ObjectZero is $\ell_\mathrm{final} = \ell_\mathrm{denoising} + \gamma\ell_\mathrm{object} \ $. 

% \begin{equation}
% \ell_\mathrm{final} = \ell_\mathrm{denoising} + \gamma\ell_\mathrm{object} \ .
% \label{eqn:final_content_loss}
% \end{equation}

Once trained, we get StyleZero and ObjectZero projectors for disentangling style and object features, respectively, from the corresponding reference images. These newly trained projectors are used in conjunction with the rest of SubZero latent modulation approach. See Appendix for training hyperparameters of StyleZero and ObjectZero.%$ are provided in Appendix.

\subsection{Extension: Zero-Order Stochastic Control}
\label{subsec:zo}
 %\vspace{-0.5 em}

Even though our method does not involve updating any parameters of the descriptor models $\psi$ and $\rho$, the optimal controller entails the need to cache intermediate activations and gradient computations as part of the chain rule, during the update of $\hat{x}_0$. Zero Order (ZO) approximation has been gaining popularity in order to alleviate the memory requirements of back-propagation. While most efforts in the context of ZO have been in the area of language modeling, we attempt to leverage ZO techniques for the latent update. To achieve zero-order optimal control, we perform our experiments by leveraging the ZO-Adam scheme described in MeZO ~\cite{malladi2024finetuninglanguagemodelsjust} and extend it to update the latent. More details and experiments are in the Appendix.

\section{Experiments}
\label{sec:exp}

\subsection{Experiment Setup}
\label{sec:expsetup}
We primarily conduct three sets of experiments: (\textit{i})~for people, we demonstrate face-style composition using single subject and style images; (\textit{ii})~we show subject-style-action composition using people and styles, while providing text prompts to perform certain actions; (\textit{iii})~finally, for common objects and pets, we conduct object-style composition.% Now, we describe the datasets as well as the metrics used throughout this work.

\begin{table*}[ht]
    %\addtolength{\tabcolsep}{-1.5pt}
    % \vspace{-0.5 em}
    \centering
    \scalebox{1}{
    \fontsize{7.0pt}{5.75pt}\selectfont
    % \fontsize{8.00pt}{8.25pt}\selectfont
    \begin{tabular}{l|c|cc|c|cc|c} 
    \toprule
    Method & Backbone & Subject Projector & Style Projector & Helper Prompts & Face Sim. & Style Sim. & Average   \\ [0.5ex] 
    \hline
    \midrule
    InstantStyle-Plus~\cite{wang2024instantstyle_plus} & \multirow{5}{*}{SDXL-Lightning} & ControlNet & IP-Adapter & & \textbf{69.0} $\pm$ 4.1  &  41.1 $\pm$ 7.7 & 55.1 \\
    InstantID~\cite{wang2024instantid} & & InstantID & ControlNet & & 54.2 $\pm$ 1.6  &  53.6 $\pm$ 4.7 & 53.9 \\
    PuLID~\cite{guo2024pulid} & & PuLID & IP-Adapter & & 56.4 $\pm$ 2.4  &  52.3 $\pm$ 4.1 & 54.4 \\
    RB-Modulation~\cite{rout2024rb} & & PuLID & StyleZero & & 59.6 $\pm$ 2.7 & 65.7 $\pm$ 4.2 & 62.7 \\
    \rowcolor{Gray} SubZero & & PuLID & StyleZero & & 64.7 $\pm$ 2.6 & \textbf{67.1} $\pm$ 4.3 & \textbf{65.9} \\
    \midrule
    InstantStyle-Plus~\cite{wang2024instantstyle_plus} & \multirow{5}{*}{SDXL-Lightning} & ControlNet & IP-Adapter & \checkmark & 65.7 $\pm$ 4.9  &  46.6 $\pm$ 9.0 & 56.2 \\
    InstantID~\cite{wang2024instantid} & & InstantID & ControlNet & \checkmark & 54.7 $\pm$ 1.6  &  63.1 $\pm$ 3.9 & 58.9 \\
    PulID~\cite{guo2024pulid} & & PuLID & IP-Adapter & \checkmark & 59.5 $\pm$ 2.1 & 58.4 $\pm$ 3.1 & 59.0 \\
    RB-Modulation~\cite{rout2024rb} &  & PuLID & StyleZero & \checkmark & 60.5 $\pm$ 1.9 & \textbf{72.7} $\pm$ 2.2 & 66.6 \\
    \rowcolor{Gray} SubZero & & PuLID & StyleZero & \checkmark & \textbf{66.5} $\pm$ 1.9 & 72.4 $\pm$ 2.4 & \textbf{69.5} \\
    \midrule
    RB-Modulation~\cite{rout2024rb} & & - & - & & 61.9 $\pm$ 1.1 & 39.3 $\pm$ 1.5 & 50.6 \\
    \rowcolor{Gray} SubZero & \multirow{-2}{*}{W\"{u}rstchen} & - & - & & \textbf{72.3} $\pm$ 1.5 & \textbf{45.5} $\pm$ 2.8 & \textbf{58.9} \\    

    \midrule
    RB-Modulation~\cite{rout2024rb} & & - & - & \checkmark & 59.7 $\pm$ 1.0 & 51.0 $\pm$ 1.5 & 55.4 \\
    \rowcolor{Gray} SubZero & \multirow{-2}{*}{W\"{u}rstchen} &  - & - & \checkmark & \textbf{69.8} $\pm$ 1.5 & \textbf{54.9} $\pm$ 2.3 & \textbf{62.3}  \\    
    \bottomrule
\end{tabular}
}
    \caption{\textbf{Face Stylization:} Results on SDXL-Lightning and W\"{u}rstchen. Helper prompts indicate the presence of style descriptions.}%We report results on SDXL-Lightning and Würstchen. Helper prompts simply mean whether the style description is present in the prompt or not.

    \label{tab:face_stylization_main}
    \vspace{-0.5 em}
\end{table*}

% \begin{table}[ht]
%     %\addtolength{\tabcolsep}{-1.5pt}
%     \vspace{1.2 em}
%     \scalebox{1}{
%     \fontsize{7.0pt}{5.75pt}\selectfont
%     \begin{tabular}{l|ccc|c}
%     \toprule
%     Method & DINO & Style Sim & HPSv2 & Average Score \\ [0.5ex] 
%     \hline
%     \midrule
%     RB-Modulation &  & & &\\
%     IP-Adapter+Pull ID &  &  & &\\
%     ZipLoRA & & & &\\
%     \rowcolor{Gray} Subzero & &  & &\\

%     \bottomrule
% \end{tabular}
%     }
%     % \fontsize{8.00pt}{8.25pt}\selectfont
%     \caption{\textbf{Results on Object +Style+Action:}}
% \end{table}

% \begin{table}[ht]
%     %\addtolength{\tabcolsep}{-1.5pt}
%     \centering
%     \fontsize{7.0pt}{5.75pt}\selectfont
%     \begin{tabular}{ccccc}
%     \toprule
%     & Style scale & Content scale & DINO Sim & CSD Sim \\ [0.5ex] 
%     \hline
%     \midrule
%     IP-Adapter & 0.2 & 0.0 & 0.241 & 0.391 \\
%     Ours & 0.2 & 0.0 & 0.241 &  0.363\\
%     IP-Adapter & 0.4  & 0.0 & 0.204 & 0.617 \\
%     Ours & 0.4 & 0.0 &  0.218 & 0.513 \\
%     IP-Adapter & 0.6  & 0.0 & 0.156 & 0.756 \\
%     Ours & 0.6 & 0.0 & 0.181 & 0.644 \\
%     IP-Adapter & 0.8  & 0.0 & 0.131 & 0.825 \\
%     Ours & 0.8 & 0.0 &  0.154 & 0.716 \\
%     IP-Adapter & 1.0  & 0.0 & 0.121 & 0.838 \\
%     Ours & 1.0 & 0.0 &  0.134 & 0.744  \\
    
%     % \rowcolor{Gray} Content & ObjectZero   & 0.610 &  - \\
%     \bottomrule
% \end{tabular}
% % \fontsize{8.00pt}{8.25pt}\selectfont
% \caption{\textbf{content and style adapters - we can remove this later.}}
% \end{table}

%\subsubsection{Datasets}
%\label{sec:datasets}
\noindent\textbf{Face Stylization Datasets.} To stylize faces, we curated a dataset consisting of 12 subjects and 30 styles. We collected a diverse range of faces across age, ethnicity and gender. Each subject provided a single image, and was asked to participate in the Human Preference Study. For stylizing the faces, we curated a dataset of 30 styles using images from StyleAligned~\cite{hertz2024style}, StyleDrop~\cite{sohn2023styledrop} and SubjectPlop~\cite{ruiz2024magic}.

% \noindent\textbf{Object-Style Composition Datasets.} For object-style composition, we use an experimental setup of ZipLoRA~\cite{shah2025ziplora}, with object images from the Dreambooth dataset~\cite{ruiz2023dreambooth}, having 30 subjects with 4-5 images per subjects. Style images are chosen from StyleDrop~\cite{sohn2023styledrop} dataset, where a single image is used per style. For each content and style, we use a single reference image.
% \noindent\textbf{Object-Style Composition Datasets.} For object-style composition, we use the experimental setup proposed by ZipLoRA~\cite{shah2025ziplora}, and select ten subjects with 4-5 images per subjects. Ten style images are chosen from StyleDrop~\cite{sohn2023styledrop} dataset, where a single image is used per style. For each content and style, we use a single reference image.
\noindent\textbf{Object-Style Composition Datasets.} For object-style composition, we use a similar setup as ZipLoRA~\cite{shah2025ziplora}, and select ten unique objects from the Dreambooth dataset~\cite{ruiz2023dreambooth}, and ten style images from StyleDrop dataset~\cite{sohn2023styledrop}.

%\subsubsection{Metrics}
%\label{sec:metrics}
\noindent\textbf{Metrics.} For object similarity we use DINO similarity score ~\cite{ruiz2023dreambooth}, i.e., cosine similarity of DINO ViT-B/6 embeddings of the object and generated images. For face similarity, we measure the cosine similarity using facial embeddings from~\cite{ye2023ip-adapter}. Further, we compute style similarity by reporting the cosine similarity between CSD embedding~\cite{somepalli2024measuring} of the reference vs. generated images. We also conduct human evaluations to quantify face stylization. For measuring performance on actions, we use the HPS-v2.1~\cite{hpsv2} score between the output image and action prompt. All metrics are computed as percentages.

\begin{figure}[h]
    \centering
    \includegraphics[width=1.0\linewidth]{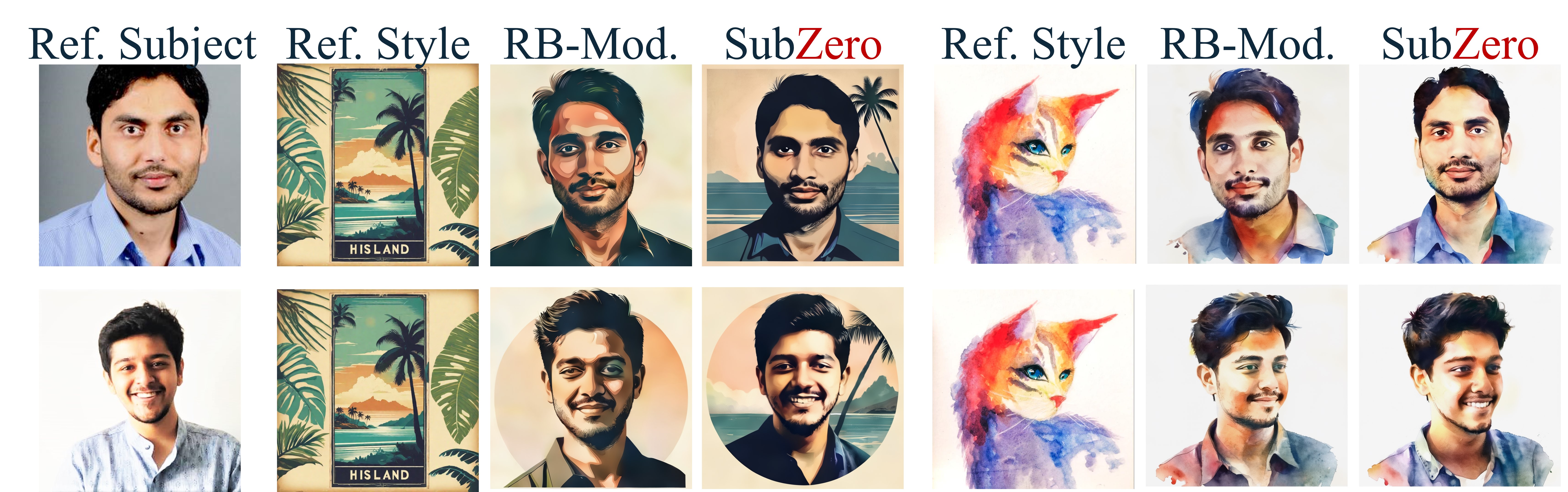}
    \vspace{- 1.5 em}
    \caption{\textbf{Comparison v/s RB-Modulation} on W\"{u}rstchen. As observed, SubZero outputs looks much more similar to the reference subject compared to RB-Modulation.}
    \vspace{- 0.7 em}
    \label{fig:wurschten_comp}
\end{figure}

\noindent\textbf{Models.}
We use two text-to-image models to achieve efficient zero-shot subject, style, and action composition: (\textit{i})~SDXL-Lightning(4-step)~\cite{lin2024sdxl} and (\textit{ii})~Stable Cascade (W\"{u}rstchen)~\cite{perniaswurstchen}. Following RB-Modulation, we use AFA-based conditioning for W\"{u}rstchen since it contains already learned CLIP-text and image projections. For experiments on SDXL-Lightning, we exploit IP-Adapters as a baseline to project the reference images to cross-attentions. For face stylization experiments with SubZero, we use PuLID as the face projector with StyleZero as the style projector. For object stylization experiments with SubZero, we use our new StyleZero and ObjectZero image projectors.

We consider several baselines for comparisons, namely, InstantStyle-Plus~\cite{wang2024instantstyle_plus}, InstantID~\cite{wang2024instantid}, RB-Modulation~\cite{rout2024rb} and Style-Aligned~\cite{hertz2024style}. Some of these baselines also exploit Controlnet~\cite{controlnet} or IP-Adapters~\cite{ye2023ip-adapter} to inject styles from reference images. All implementation details and hyperparameters are provided in the Appendix.

% \paragraph{Latent Modulation Details.}
% We run stochastic optimal control on the latent for $M=5$ iterations during each denoising step for both Stable-Cascade and SDXL-Lightning

\subsection{Results}\label{sec:results}
%We now present our extensive evaluations to demonstrate the effectiveness of our proposed SubZero framework. %and StyleZero/ObjectZero projectors.

\subsubsection{Face Style Composition}\label{sec:faceRes}
%For the face stylization tasks, we encode the style reference image using our StyleZero projector and face images using PuLID~\cite{guo2024pulid}. Then, we perform the latent modulation using our proposed SubZero objectives and orthogonal temporal aggregation of various cross-attentions among text, style, and face modalities. 
As observed in Fig.~\ref{fig:faces_stylized}, SubZero can effectively stylize the given faces into a diverse range of styles.% very effectively. 

\noindent\textbf{Quantitative Comparisons.} We compare SubZero against several state-of-the-art tuning-free personalization methods for SDXL-Lightning and W\"{u}rstchen architectures, with and without ``helper prompts'' (i.e., whether or not style description is present in the text prompt). We provide mean scores over 3 random seeds. Table~\ref{tab:face_stylization_main} presents our main result: SubZero produces the best images for personal (face)-similarity and style-similarity with or without helper prompts. For instance, while InstantStyle-Plus~\cite{wang2024instantstyle_plus} achieves higher face-similarity score for SDXL-Lightning without helper prompts, it achieves significantly lower style-similarity than our proposed technique. This suggests that while InstantStyle-Plus is good at reproducing faces due to ControlNet, it performs suboptimal stylization. Similarly, while RB-Modulation~\cite{rout2024rb} achieves good stylization for SDXL-Lightning with helper prompts, it cannot capture faces accurately. SubZero significantly outperforms the prior art as it achieves the highest average similarity score and establishes a new state-of-the-art for face stylization.

\begin{figure}[h]
    \centering
    \includegraphics[width=0.9\linewidth]{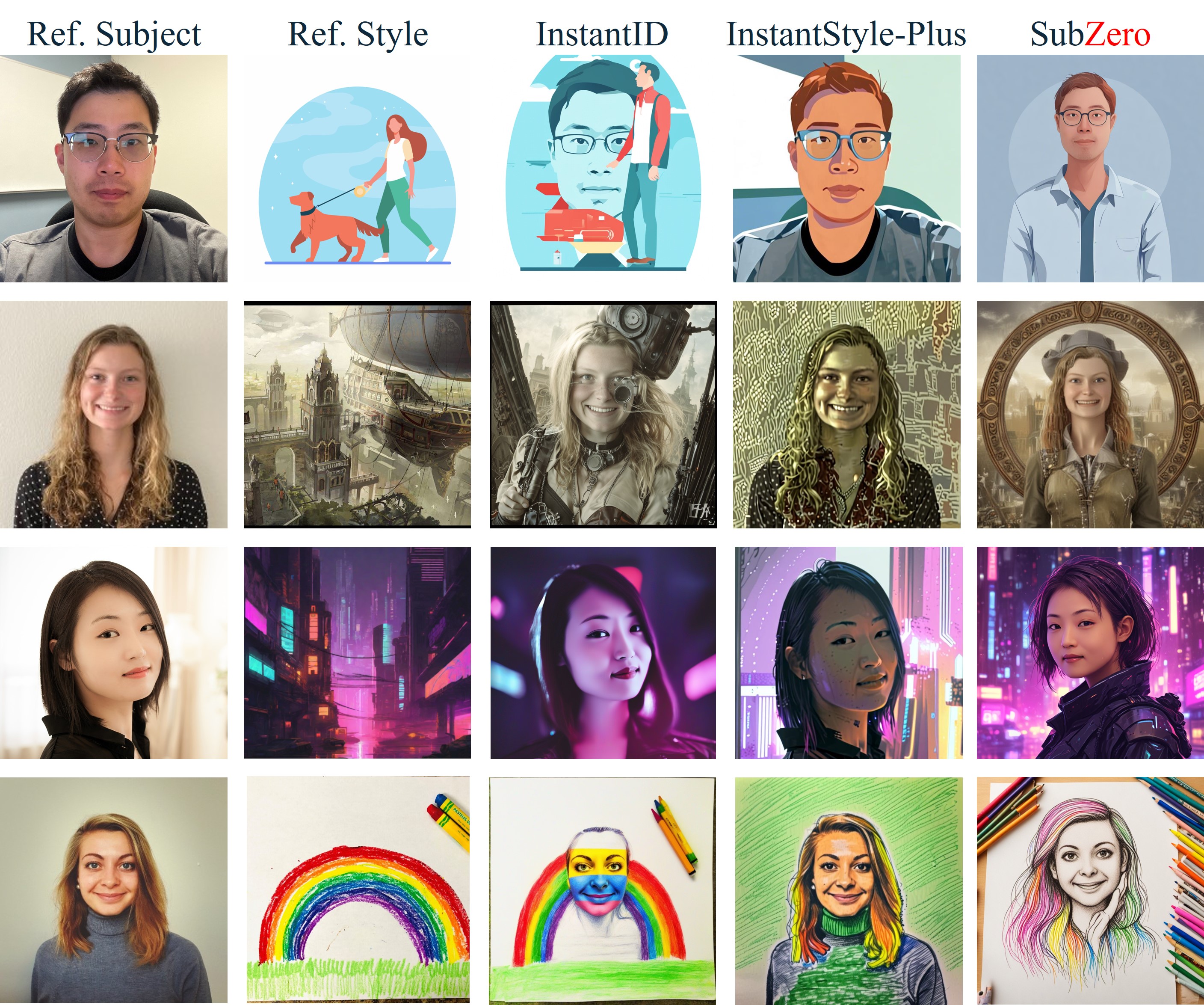}
    %\vspace{- 1.2 em}
    \caption{\textbf{Visual comparison} between SubZero and ControlNet/DDIM Inversion based schemes. SubZero is more flexible and reduces subject leakage.} 
    \label{fig:vs_controlnet_schemes}
    \vspace{- 0.6 em}
\end{figure}

\noindent\textbf{Qualitative Comparisons.} Next, we compare SubZero and RB-Modulation~\cite{rout2024rb} in Fig.~\ref{fig:wurschten_comp}. As evident, SubZero is significantly more effective at maintaining the correct subject through various styles. In contrast, RB-Modulation fails to preserve the correct face while performing stylization. 
In Fig.~\ref{fig:vs_controlnet_schemes}, we compare against InstantX methods~\cite{wang2024instantid, wang2024instantstyle_plus} that employ ControlNet and/or DDIM-inversion for subject-style composition. As observed, InstantID often leaks irrelevant content from style reference into the final generated image or suffers from undesirable artifacts. On the other hand, InstantStyle-Plus achieves good stylization but it is too rigid due to ControlNet; this results in significantly less diverse output images. Clearly, SubZero outperforms these methods in both diversity as well as stylization quality.

\noindent \textbf{Human Preference Study:} We surveyed 10 subjects who provided their photos, by using a customized human evaluation form \textit{containing their own images}, as shown in the Appendix. Each form had three sections, the results of which are summarized in Table~\ref{tab:human}. Each section had 10 styles. Hence, our evaluation contains 300 responses. We place generated images from various models side-by-side v/s subzero and ask humans to pick the image which most resembles their face while best aligning with the reference style image. As observed in Table~\ref{tab:human}, SubZero was the preferred choice at \textbf{64.1}$\%$ v/s the PuLID+IP-Adapter baseline,  \textbf{64.5}$\%$ v/s RB-Modulation(on W\"{u}rstchen) and \textbf{74.7}$\%$ v/s InstantStyle by the human subjects themselves.

\begin{table}[ht]
    %\addtolength{\tabcolsep}{-1.5pt}
    \centering
    \vspace{-0.5 em}

    \fontsize{7.0pt}{5.75pt}\selectfont
    \begin{tabular}{l|c|c|c}
    \toprule
    Method & v/s PuLID+IP-Adapter & v/s InstantStyle-Plus & v/s RB-Mod \\ [0.5ex] 
    \hline
    \midrule
    Not Subzero & 21.7 & 24.0 & 11.8 \\
    Tie & 14.1 & 1.3 & 23.7  \\
    \rowcolor{Gray} SubZero & \textbf{64.1}  & \textbf{74.7} & \textbf{64.5} \\
    \bottomrule
\end{tabular}
% \fontsize{8.00pt}{8.25pt}\selectfont

\caption{\textbf{Human Evaluation for Face Stylization.}}
 \vspace{-1.2 em}
\label{tab:human}
\end{table}

\vspace{-0.5 em}
\begin{figure}[h]
    \centering
    \includegraphics[width=0.9\linewidth]{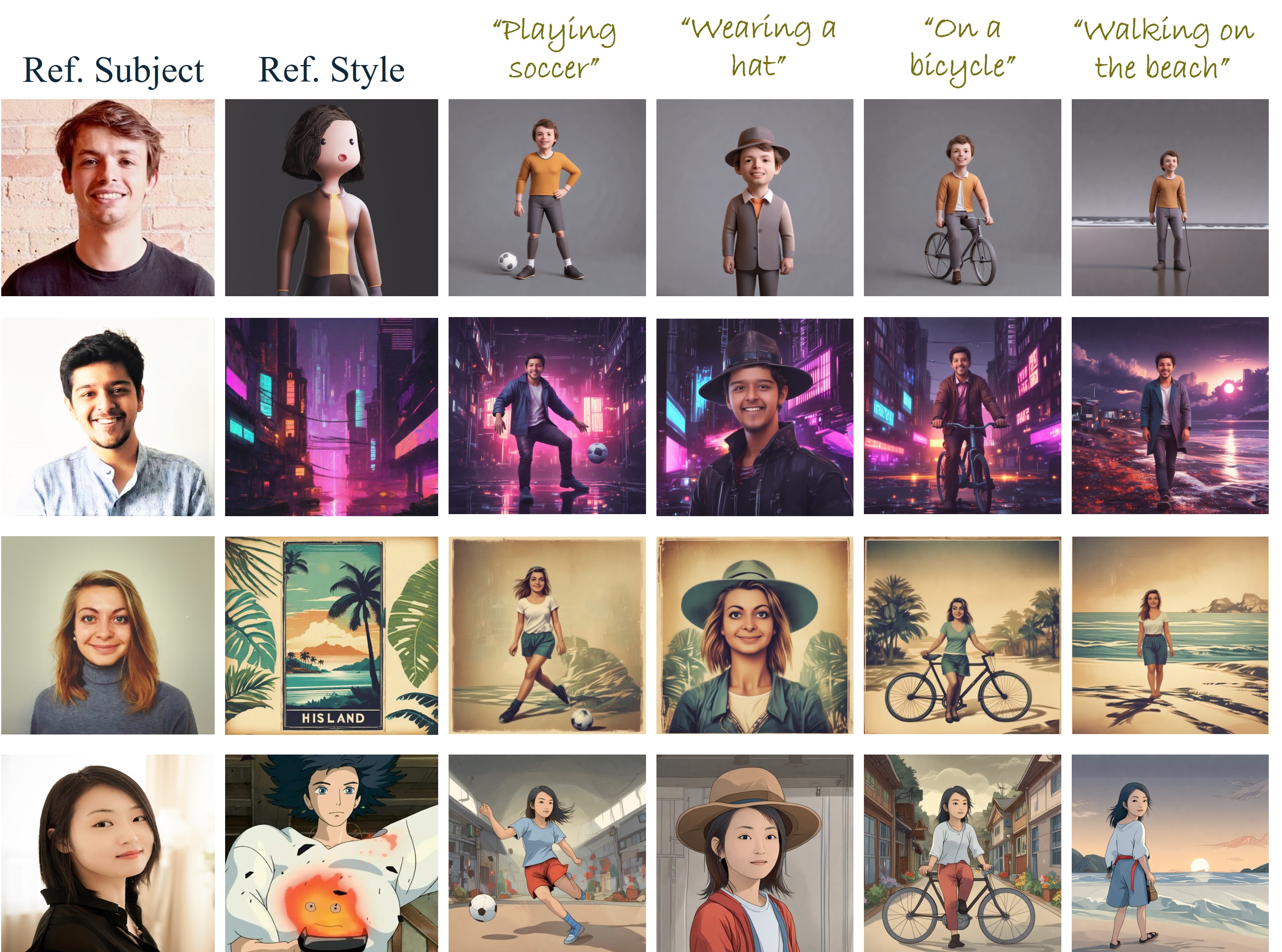}
    %\vspace{- 1.2 em}
    \caption{\textbf{Face, Style and Action composition} using SubZero.}
    \vspace{- 0.8 em}
    \label{fig:face_style_action}
    
\end{figure}

\subsubsection{Face-Style-Action Composition}
Could we compose the face of any subject in any style performing any action in a zero-shot setting? 
We explore this aspect using SubZero and evaluate it on face stylization for a set of actions described by action prompts. Table~\ref{tab:face_style_action} shows the results across 12 subjects, 10 Actions and 10 Styles and an average across 3 seeds. We report the Human Preference Scores (HPSv2), in addition to the usual face- and style-similarities. We notice that SubZero improves significantly over the baselines especially on the HPSv2 score. RB Modulation suffers from content style leakage through AFA which makes it harder to generate more diverse images. Since SubZero exploits our proposed orthogonal temporal aggregation strategy for the cross-attentions across multiple modalities, we achieve significantly stronger results. Additionally, ControlNet and DDIM inversion prove to hinder flexibility, resulting in lower HPSv2 scores for InstantX based methods. 
Our results can be visualized in Fig.~\ref{fig:face_style_action}. 
\label{sec:facestyleactionRes}

% \begin{figure}[t!]
%     \centering
%     \renewcommand{\tabcolsep}{1pt}
%     \newcommand{\mywidth}{0.5\linewidth}
%     \resizebox{\linewidth}{!}{% take the[] entire width, but still find a good per-image width otherwise text gets compressed
%     \begin{tabular}{cccc}
%     Content & Style & IP-Adapter & Ours \\
%     \includegraphics[width=\mywidth]{figures/object_style_composition/content/candle_content.jpg}
%     & \includegraphics[width=\mywidth]{figures/object_style_composition/style/image_01_22.jpg}
%     & \includegraphics[width=\mywidth]{figures/object_style_composition/ip/prompt_A_candle_content_candle_0.4_style_sty_22_0.6.png} & \includegraphics[width=\mywidth]{figures/object_style_composition/ours/prompt_A_candle_content_candle_0.4_style_sty_22_0.6.png}
%     \\
%     \end{tabular}}
%     \caption[]{Caption here.}
%     \label{fig:object_style_action}
% \end{figure}

\begin{table}[ht]
    %\addtolength{\tabcolsep}{-1.5pt}
    \vspace{-0.5 em}
    \scalebox{0.95}{
    \fontsize{7.0pt}{5.75pt}\selectfont
    \begin{tabular}{l|ccc|c}
    \toprule
    Method & Face Sim. & Style Sim. & HPSv2 & Average \\ [0.5ex] 
    \hline
    \midrule
    InstantStyle-Plus~\cite{wang2024instantstyle_plus} & 66.0 & 47.3 & 24.6 & 46.0 \\
    InstantID~\cite{wang2024instantid} & 62.3 & 58.2 & 22.4 & 47.6 \\ 
    PulID+IP-Adapter~\cite{guo2024pulid} & 58.9 & 56.0 & 24.9 & 45.9 \\
    RB-Modulation~\cite{rout2024rb} & 58.3 & 72.6 & 24.8 & 51.9 \\
    \rowcolor{Gray} SubZero & \textbf{64.2} & \textbf{73.1} & \textbf{26.1} & \textbf{54.5} \\

    \bottomrule
\end{tabular}
    }
    % \fontsize{8.00pt}{8.25pt}\selectfont
    \caption{\textbf{Results on Face+Style+Action:} We report results using SDXL-Lightning as a backbone and compare SubZero against SOTA methods for composing subjects, styles and actions.}
    \vspace{- 1.5 em}

    \label{tab:face_style_action}
\end{table}

\begin{figure}[ht]
    \centering
    \includegraphics[width=0.85\linewidth]{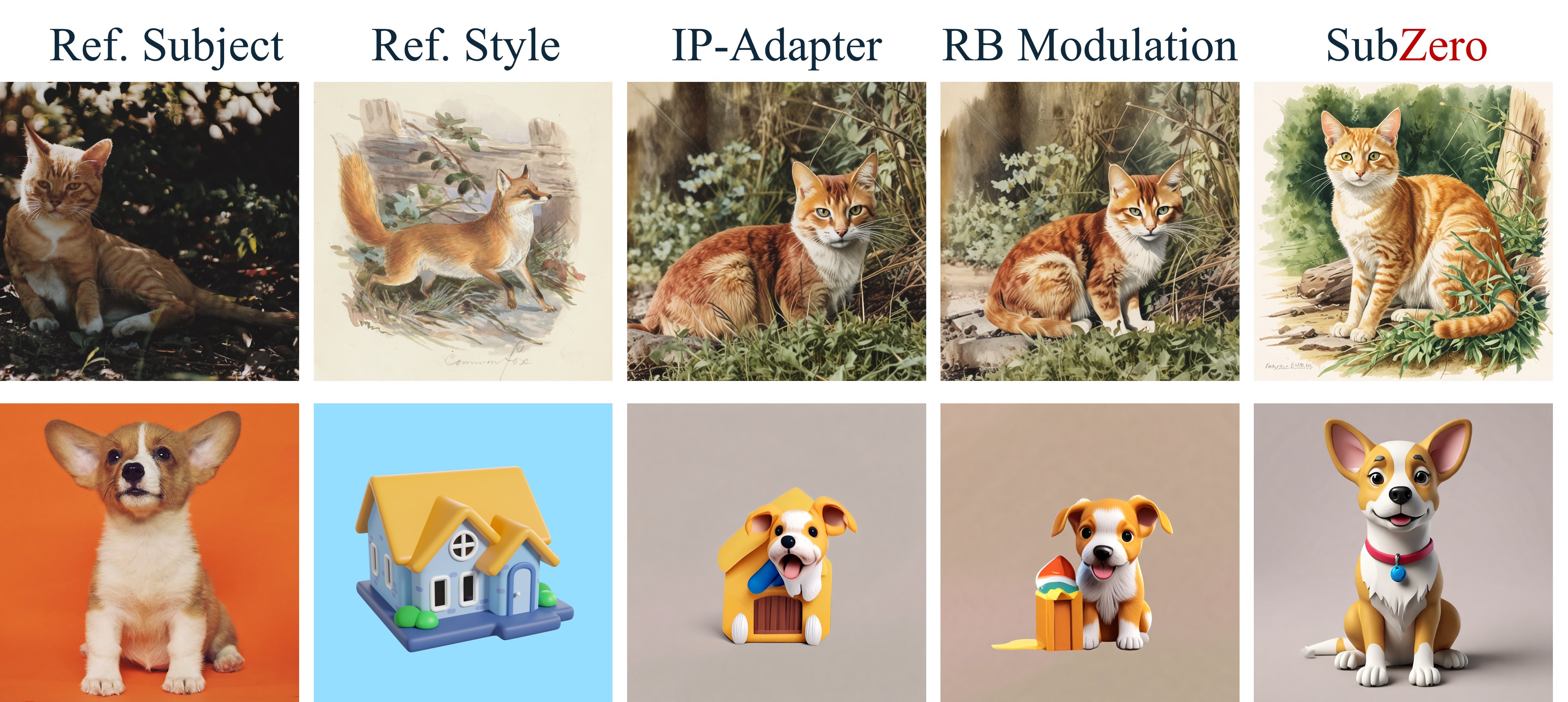}
    \caption{\textbf{Object and Style composition} using SubZero.}
    \vspace{- 1 em}
    \label{fig:object_style_action}
\end{figure}

\subsubsection{Object-Style Composition}\label{sec:objstyleRes}
We now evaluate the ability of SubZero to compose any object in any style in a zero-shot manner using our newly trained StyleZero and ObjectZero projectors. To this end, we use all subjects from the DreamBooth dataset and 20 styles from StyleDrop~\cite{sohn2023styledrop} to perform object-style composition for 600 object-style pairs. %We select the following subjects to show the effectiveness of our approach: []. 
Table~\ref{tab:obj_stylization_main} shows we achieve a very high DINO score, demonstrating the strong ability of SubZero to maintain the correct content while generating zero-shot stylized images. On SDXL-Lightning, we also achieve the best style similarity. On an average, we significantly outperform the IP-Adapter, RB-Modulation and StyleAligned baselines. 

Fig.~\ref{fig:object_style_action} shows the qualitative comparison between IP-Adapter, RB-Modulation, and SubZero. Notably, both IP-Adapter and RB-Modulation show irrelevant content leakage (e.g., see the house/hut structure getting leaked into the bottom dog). In contrast, SubZero performs the object-style composition without any leakage. This clearly highlights the superiority of SubZero compared to existing methods.

\begin{table}[ht]
    \addtolength{\tabcolsep}{-1.5pt}
    \vspace{-0.5 em}
    \centering
    \scalebox{1}{
    \fontsize{7.0pt}{5.75pt}\selectfont
    % \fontsize{8.00pt}{8.25pt}\selectfont
    \begin{tabular}{l|c|cc|c} 
    \toprule
    Method & Backbone & DINO Sim. & Style Sim. & Average \\ [0.5ex] 
    \hline
    \midrule
    % ZipLoRA~\cite{shah2025ziplora} & SDXL & 52.2 & 33.8 & 43.0\\
    StyleAligned~\cite{hertz2024style} & SDXL & 36.8 & 51.0 & 43.9 \\

    \midrule
    IP-Adapter~\cite{ye2023ip-adapter} & \multirow{3}{*}{Lightning} & 46.0 & 36.2 & 41.1 \\
    RB-Mod~\cite{rout2024rb}+IP-Apapter &  & 48.7 & 58.8 & 53.8 \\
    \rowcolor{Gray} SubZero & & \textbf{53.5} & \textbf{61.4} & \textbf{57.5} \\
    \midrule
    RB-Modulation~\cite{rout2024rb} & & 42.6  & \textbf{44.2} & 43.4 \\
    \rowcolor{Gray} SubZero & \multirow{-2}{*}{W\"{u}rstchen} & \textbf{63.2} & 44.0 & \textbf{53.6} \\    
    \bottomrule
\end{tabular}
}
    \vspace{- 0.5 em}
    \caption{\textbf{Object-Style Composition:} We report results on SDXL-Lightning and W\"{u}rstchen and compare SubZero against IP-Adapter, Style-aligned and RB-Modulation.} 

    \label{tab:obj_stylization_main}
\end{table}

\begin{table}[h]
    %\addtolength{\tabcolsep}{-1.5pt}

    \centering
    \fontsize{7.0pt}{5.75pt}\selectfont
    \begin{tabular}{c|cc|ccc}
    \toprule
    Helper Prompt & Dis. Control & OTA & Face Sim. & Style Sim. & Average\\
    \hline
    \midrule
       &  & & 57.7 & 54.1 & 55.9 \\
       \rowcolor{Gray} &  & \checkmark & 59.0 & 53.4 & 56.2 \\
       \rowcolor{Gray} & \checkmark & \checkmark & \textbf{64.7} & \textbf{67.1} & \textbf{65.9} \\
      \hline
     \checkmark & & & 59.5 & 58.4 & 59.0 \\
      \rowcolor{Gray} \checkmark & & \checkmark & 60.1 & 61.9 & 61.0 \\
      \rowcolor{Gray} \checkmark & \checkmark & \checkmark & \textbf{66.5} & \textbf{72.4} & \textbf{69.5} \\
    \bottomrule
\end{tabular}
% \fontsize{8.00pt}{8.25pt}\selectfont
\vspace{-0.7 em}
\caption{\textbf{Individual gain from SubZero components:} We report results on SDXL-Lightning with StyleZero and PulID.}\label{table:face_ablation}
\vspace{-1.0 em}

\end{table}

\subsection{Ablation Studies}\label{sec:ablations}
\noindent\textbf{Individual gain from all inference components.} Table \ref{table:face_ablation} shows the individual gain from our proposed Disentangled Latent Optimization~\ref{subsec:dsoc} and the Orthogonal Temporal Aggregation (OTA) scheme, both with and without helper prompts. We perform this experiment on the face stylization task from Table~\ref{tab:face_stylization_main}. Results are on an SDXL-Lightning baseline, with PuLID as the subject projector and StyleZero as the style projector. As observed, OTA improves the Average score by \textbf{0.3} to \textbf{2}$\%$, and the latent optimizer further improves the it by \textbf{9.5}$\%$. Overall, both the methods compliment each other and contribute significant gains.

\noindent\textbf{Impact of Style Projectors.} We demonstrate the effectiveness of our StyleZero projector compared to existing style projectors IP-Adapter~\cite{ye2023ip-adapter} and StyleCrafter~\cite{liu2023stylecrafter}, on face style composition in Table~\ref{table:stylezero_faces}. As observed, SubZero works standalone with all existing style projectors and with StyleZero we observe a \textbf{1.4} to \textbf{1.8}$\%$ improvement.

\begin{table}[h]
    %\addtolength{\tabcolsep}{-1.5pt}

    \vspace{-0.3 em}
    \centering
    \fontsize{7.0pt}{5.75pt}\selectfont
    \begin{tabular}{lc|ccc}
    \toprule
    Method & Style Projector & Face Sim & Style Sim & Average\\
    \hline
    \midrule
      \rowcolor{Gray}  & IP-Adapter~\cite{ye2023ip-adapter} & 65.9 & 70.3 & 68.1 \\
      \rowcolor{Gray}  & StyleCrafter~\cite{hertz2024style} & 63.5 & 71.9 & 67.7 \\
     \rowcolor{Gray}  \multirow{-3}{*}{SubZero} & StyleZero & \textbf{66.5} & \textbf{72.4} & \textbf{69.5} \\
    \bottomrule
\end{tabular}
% \fontsize{8.00pt}{8.25pt}\selectfont
\vspace{-0.5 em}
\caption{\textbf{SubZero with various facial style projectors.}}\label{table:stylezero_faces}  
 \vspace{-0.7 em}
\end{table}

\vspace{-0.9 em}
\section{Conclusion}\label{sec:conc}\vspace{-2mm}

In this paper, we proposed SubZero, which is a framework for robust and efficient zero-shot face, style and action composition. This consists of a Disentangled Stochastic Optimal Controller to inject subjects and styles into latents without causing any leakage. It also consists of the Orthogonal Temporal Aggregation scheme for Cross-Attention features originating from subject, style and text conditioning. We further proposed a novel method to train customized content and style projectors to reduce content and style leakage. Additionally, we discuss the feasibility of Zero-Order optimization for performing Stochastic Optimal Control. Through extensive experiments, we show that SubZero can significantly improve performance over the current state-of-the-art. Our proposed approach is suitable for running on-edge, and shows significant improvements over previous works performing subject, style and action composition. Assessing the performance of SubZero, we believe that our proposed method will lay a foundation for further research in training-free personalization.

{
    \small    
    \bibliographystyle{ieeenat_fullname}
    \bibliography{main}
}

\clearpage
\onecolumn
\setcounter{page}{1}
\appendix
\appendixpage
\counterwithin{figure}{section}
\counterwithin{table}{section}
%-------------------------------------------------
\section{Contents}
\label{sec:SuppleIntro}
As part of the supplementary materials for this paper, we share our Implementation details and show extended qualitative and quantitative results for our proposed approach. The supplementary materials contain: 

\begin{easylist}[itemize]
@ Datasets 
@ Implementation Details and Hyperparameters
@ Quantitative Results
    @@ Standalone StyleZero and ObjectZero adapters
    @@ Varying style and content scaling
    @@ Subject leakage 
    measurement
    @@ Runtime analysis
    @@ Zero-order stochastic optimal control    
@ Qualitative Results
    @@ Face style composition
        @@@ With style helper prompts
        @@@ Without style helper prompts
    @@ Object style composition
@ Limitations and Future Work
\end{easylist}

\section{Datasets}
\label{subsec:data_app}
\begin{figure}[h]
    \centering
    \includegraphics[width=1.0\linewidth]{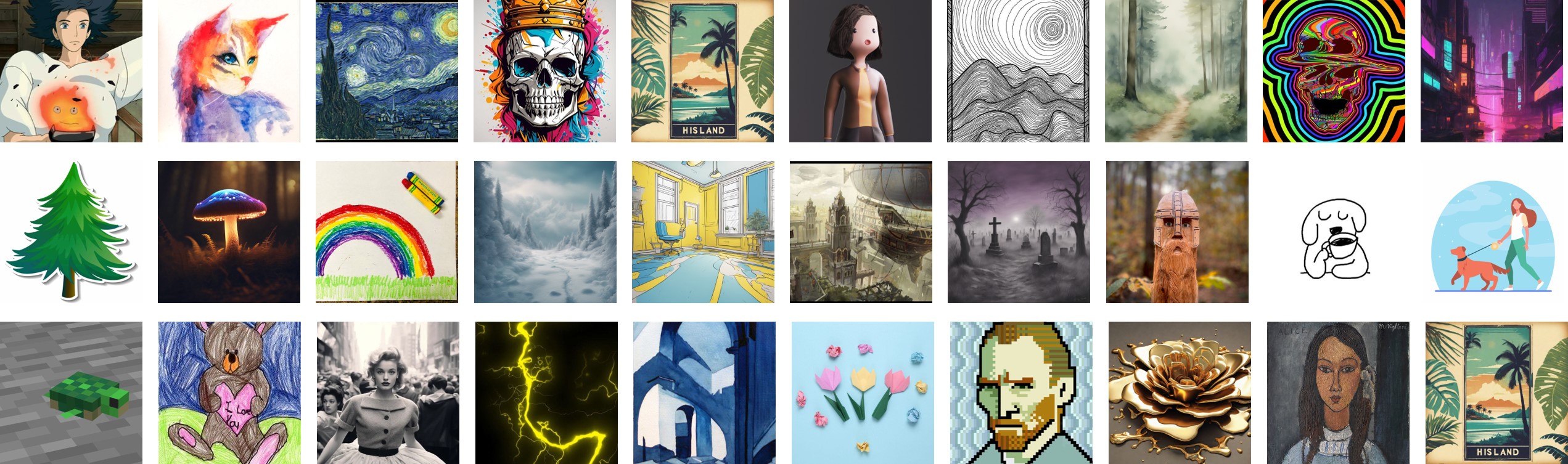}
    \caption{
    % \textbf{Stylized Faces with SubZero}. Stylized Faces with SubZero.
All the style images from our face-style composition dataset
    }%, and suffer from mode collapse. However, our proposed FouRA produces more diverse images.}
    \label{fig:all_styles}

\end{figure}

\paragraph{Face-Style Composition.} As discussed in Section~\ref{sec:expsetup}, we curate a dataset with 12 faces \textbf{which remain unseen by our foundational models}. We do not use a public dataset, as we observe that celebrity faces and AI generated faces are easy for foundational models to replicate, as these faces might have been seen before. Hence, we collect our own dataset, with faces which are not seen before. The images shared with us are directly by the subjects themselves. Moreover, each subject is invited to participate in our user study in Table~\ref{tab:human}. Of the 12 subjects, 10 participated in the study. For styles, we collect 30 vivid styles from datasets such as SubjectPlop~\cite{ruiz2024magic}, StyleDrop~\cite{sohn2023styledrop} and StyleAligned~\cite{hertz2024style}. All style images are shown in Figure~\ref{fig:all_styles}. For each result in Tables~\ref{tab:face_stylization_main},\ref{table:face_ablation},\ref{table:stylezero_faces},\ref{table:latency}, we perform analysis over 12 subjects, 30 styles and 3 seeds, totaling \textbf{1080 samples}.

\paragraph{Face-Style-Action Composition.} As discussed in Section~\ref{sec:facestyleactionRes}, we use a dataset with 12 faces, 10 styles and 10 action prompts over 3 seeds for action generation. This totals inference over \textbf{3600 samples}.  We list the 10 action prompts below.
\begin{python}
1. wearing a jacket
2. walking on the beach
3. laughing
4. playing soccer
5. dancing
6. punching 
7. on a bicycle
8. wearing a hat
9. holding a mike
10. giving a speech to an audience
\end{python}

\begin{figure}[h]
    \centering
    \includegraphics[width=0.8\linewidth]{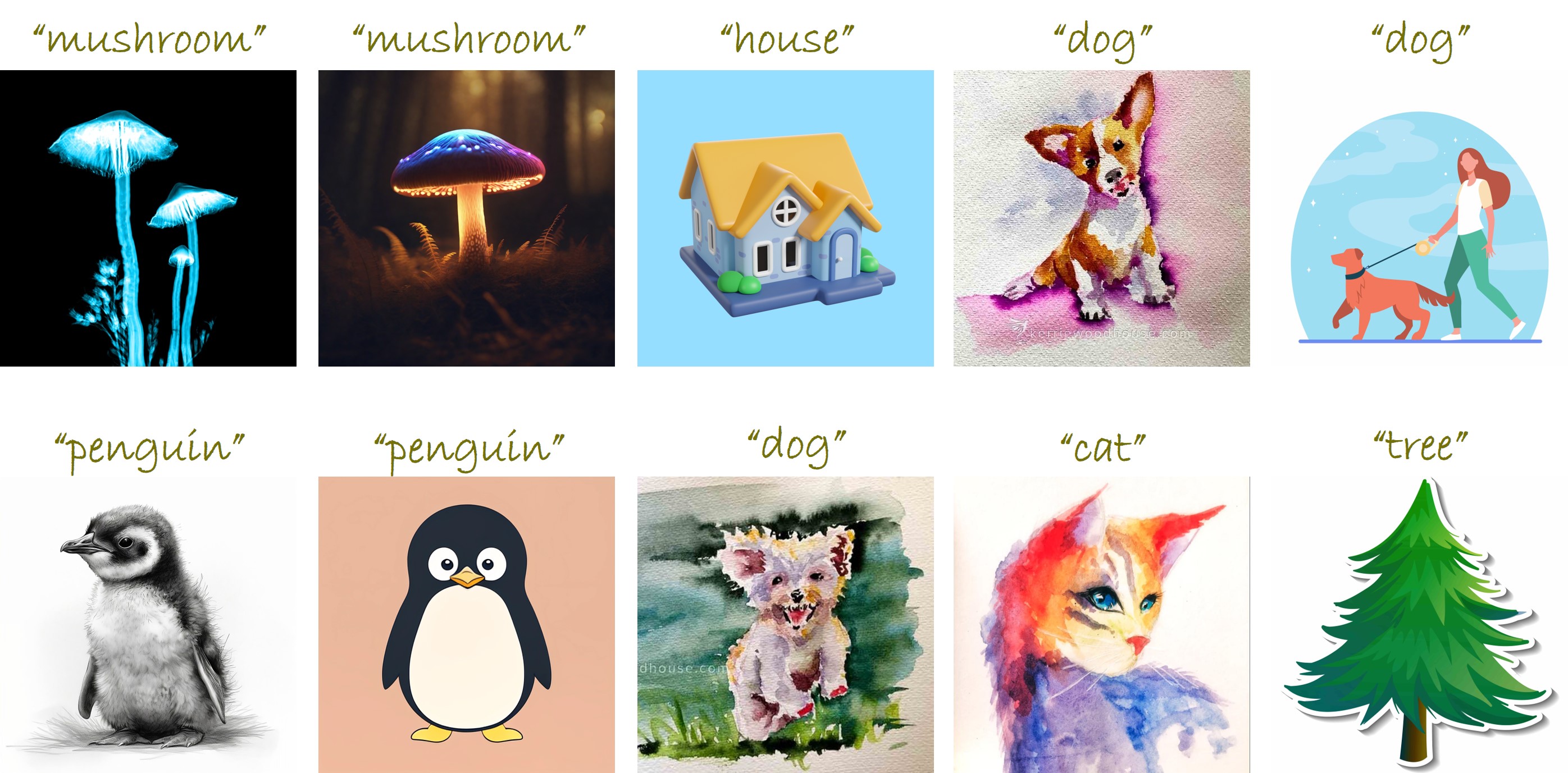}
    \caption{
    % \textbf{Stylized Faces with SubZero}. Stylized Faces with SubZero.
All the style images from our style leakage dataset, along with leakage prompts
    }%, and suffer from mode collapse. However, our proposed FouRA produces more diverse images.}
    \label{fig:leakage_dataset}

\end{figure}

\paragraph{Subject Leakage.}
To measure the subject leakage problem in further detail, we curate a dataset of 10 styles, each of which contain a salient object. These images are shown in Figure~\ref{fig:leakage_dataset}. To measure leakage along with Style Similarity, we compute the CLIP-ViT-L distance between the generated images and "leakage prompts" which describe the salient subject in the style image. This analysis is further detailed in Section~\ref{subsec:leakage}.

\paragraph{Object-Style Composition}
We use a set of ten unique subject images from the Dreambooth dataset~\cite{ruiz2023dreambooth}, and visualize them in figure~\ref{fig:content_images}. In addition, we select ten unique style images from the StyleDrop dataset~\cite{sohn2023styledrop}, shown in figure~\ref{fig:style_images}. We run inference over 3 seeds. Hence, object stylization results are over \textbf{300 samples}.

\begingroup
\setlength{\tabcolsep}{2pt} % Default value: 6pt
\renewcommand{\arraystretch}{0.7} % Default value: 1

\begin{figure}
    \centering
    \begin{tabular}{ccccc}
    \includegraphics[width=0.15\linewidth]{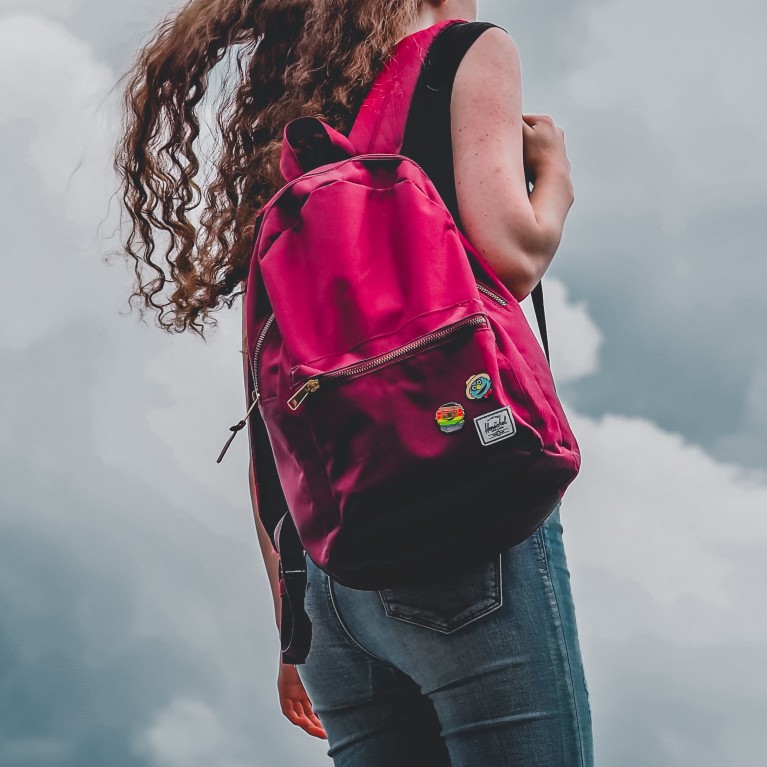} & \includegraphics[width=0.15\linewidth]{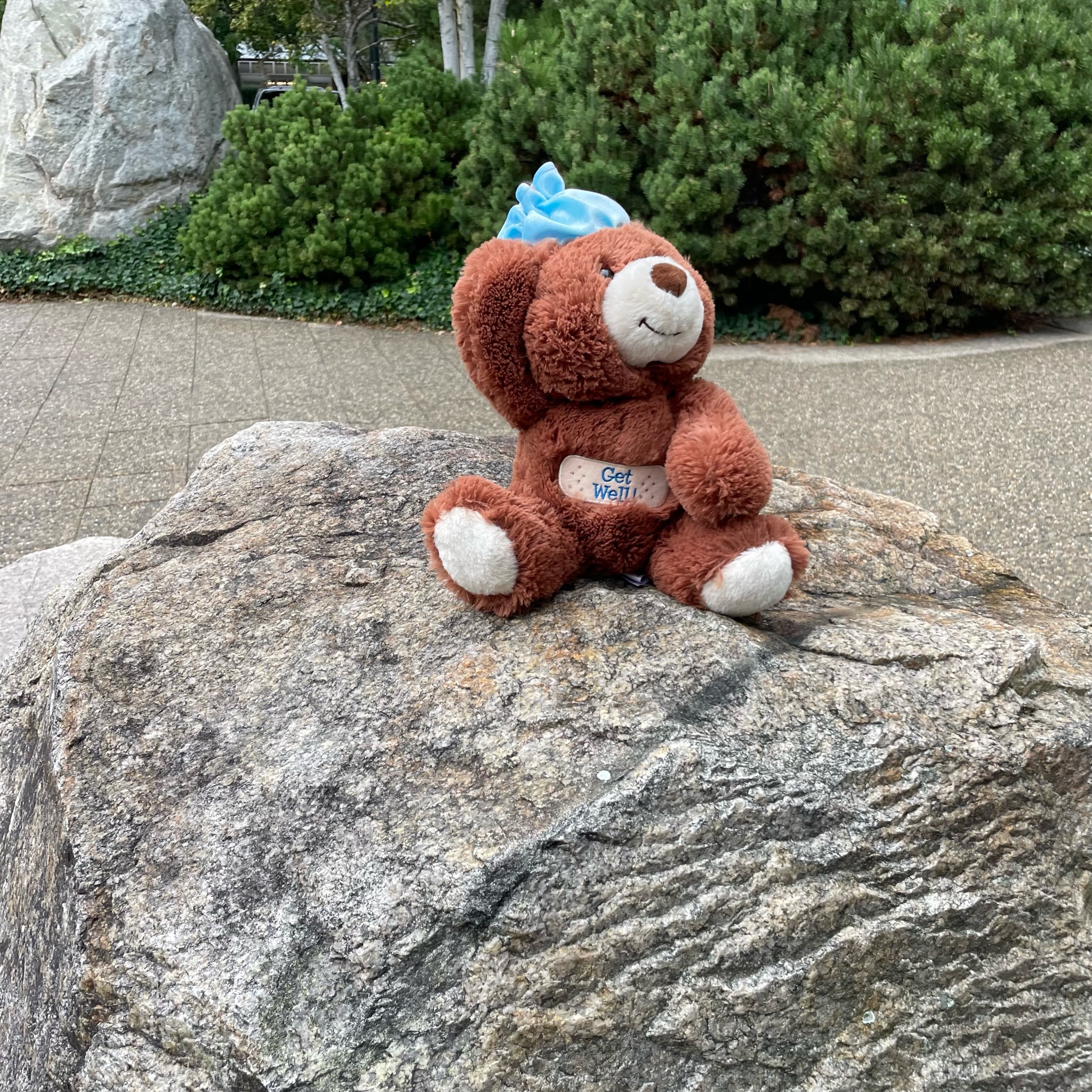} &  \includegraphics[width=0.15\linewidth]{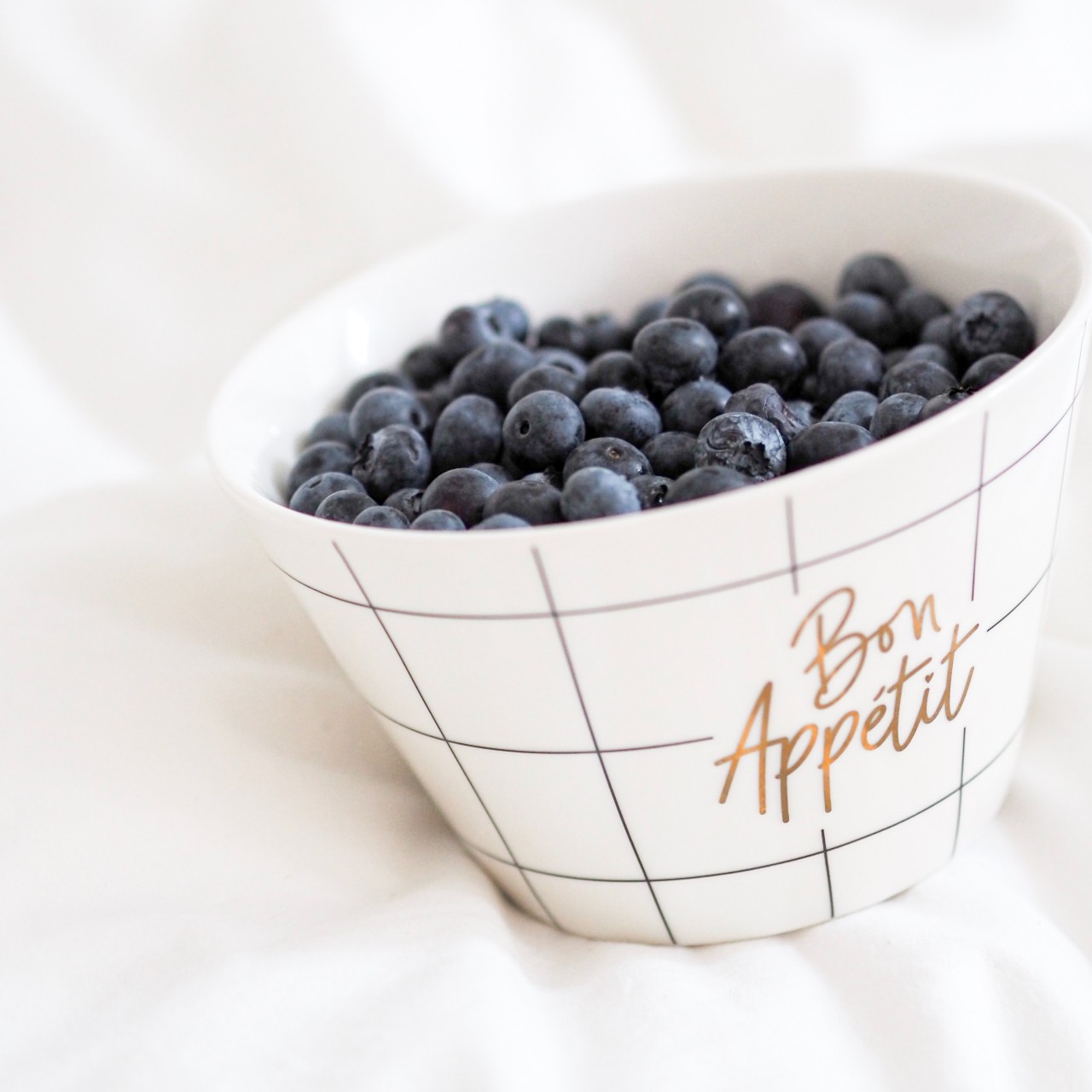} &
     \includegraphics[width=0.15\linewidth]{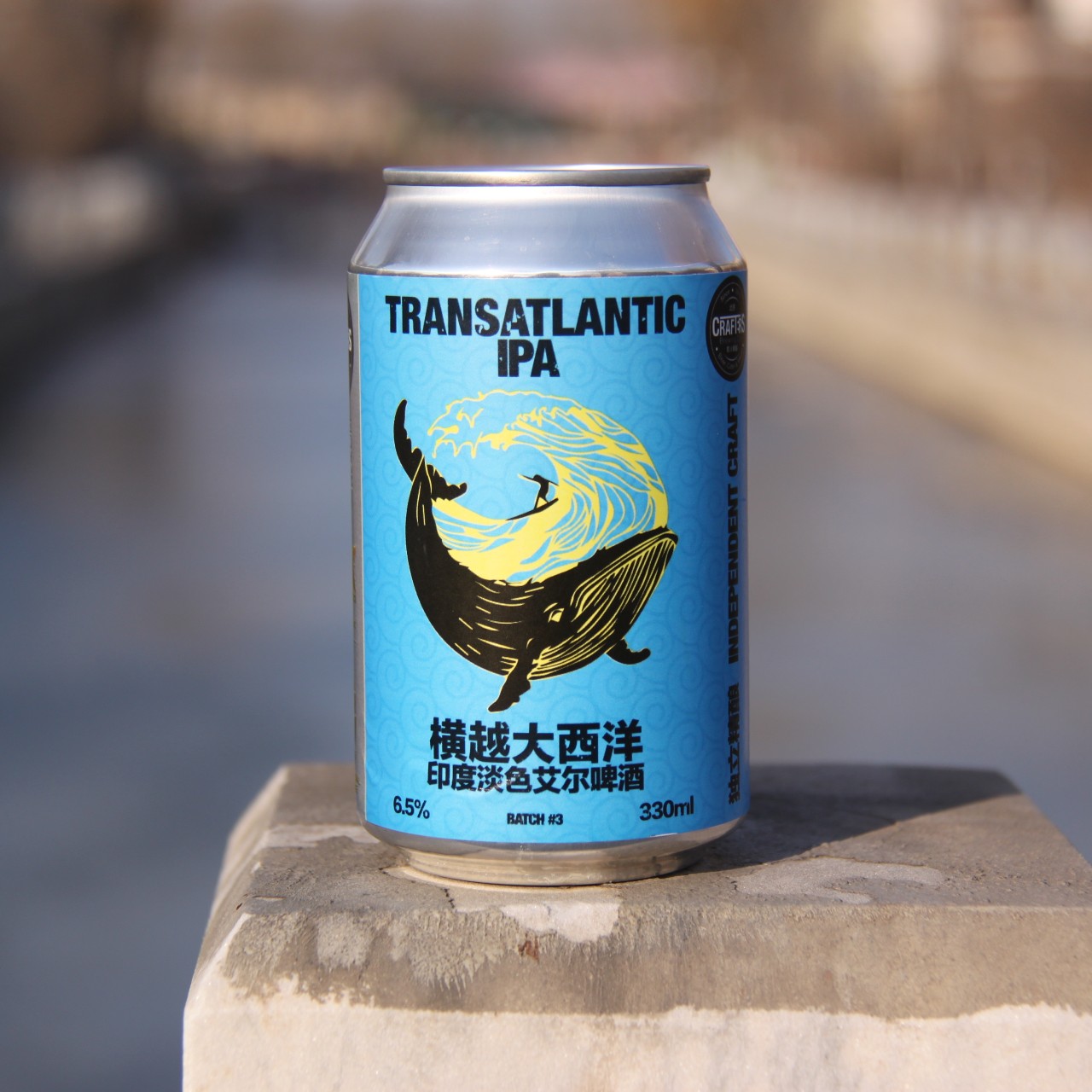} &
      \includegraphics[width=0.15\linewidth]{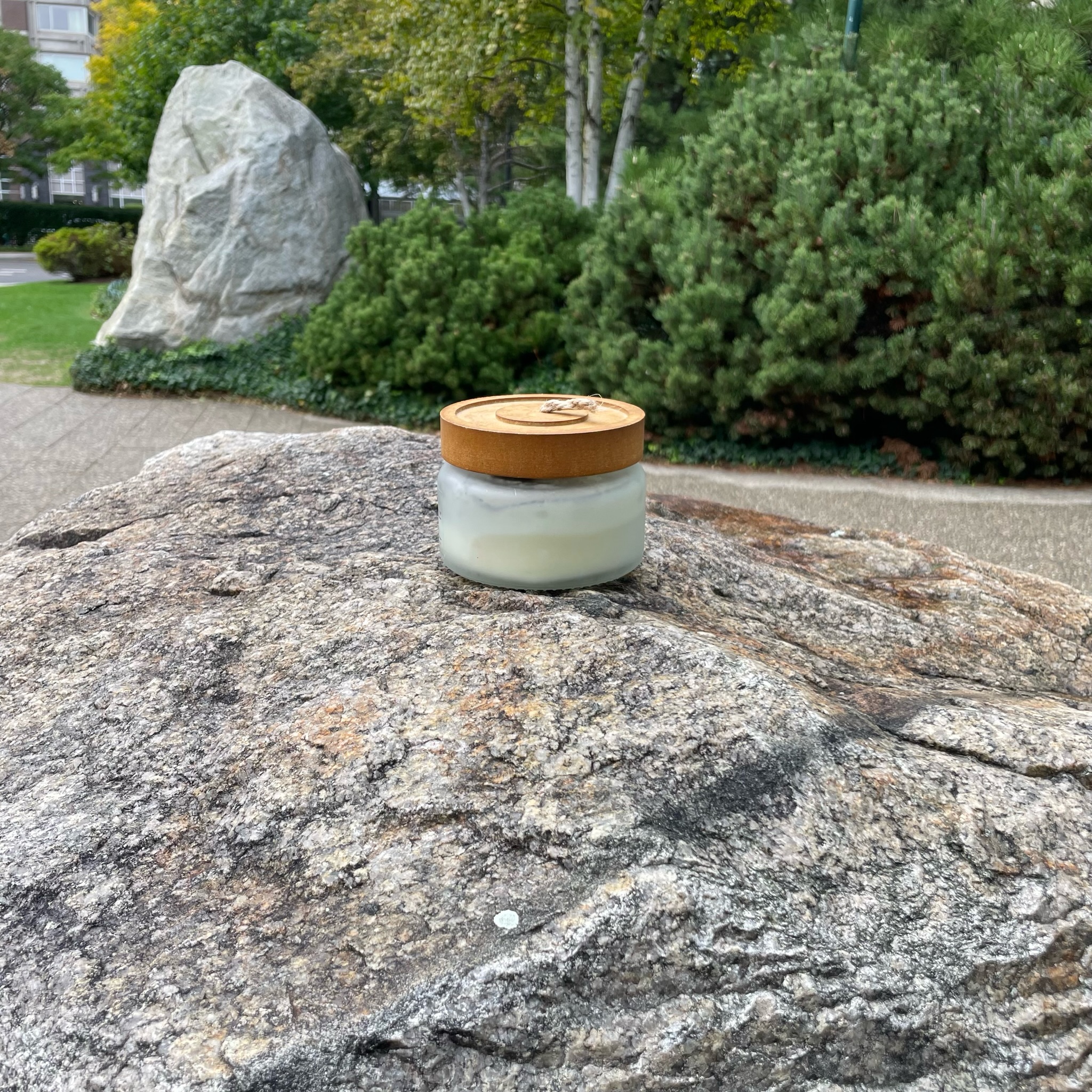} \\
       \includegraphics[width=0.15\linewidth]{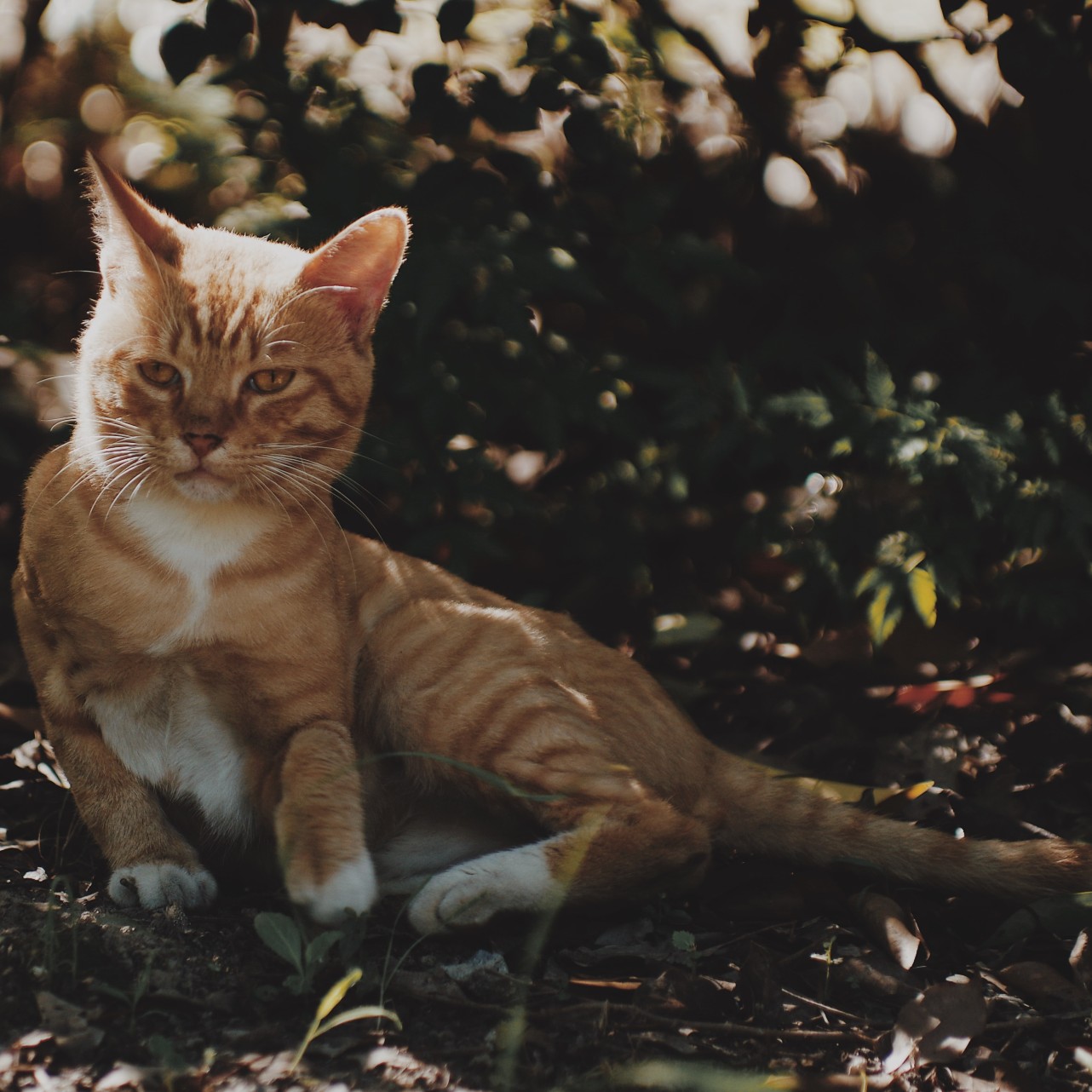} & \includegraphics[width=0.15\linewidth]{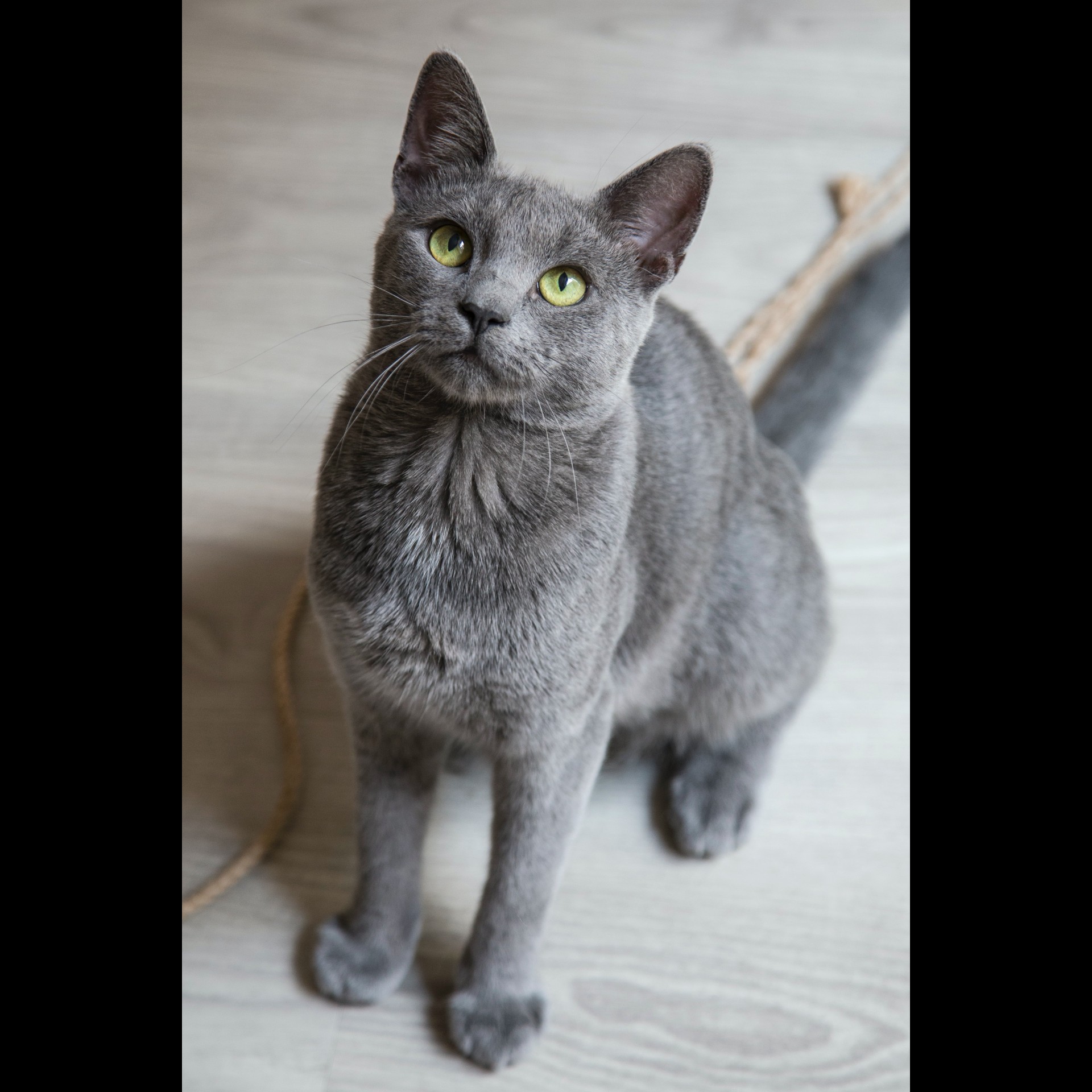} & 
       \includegraphics[width=0.15\linewidth]{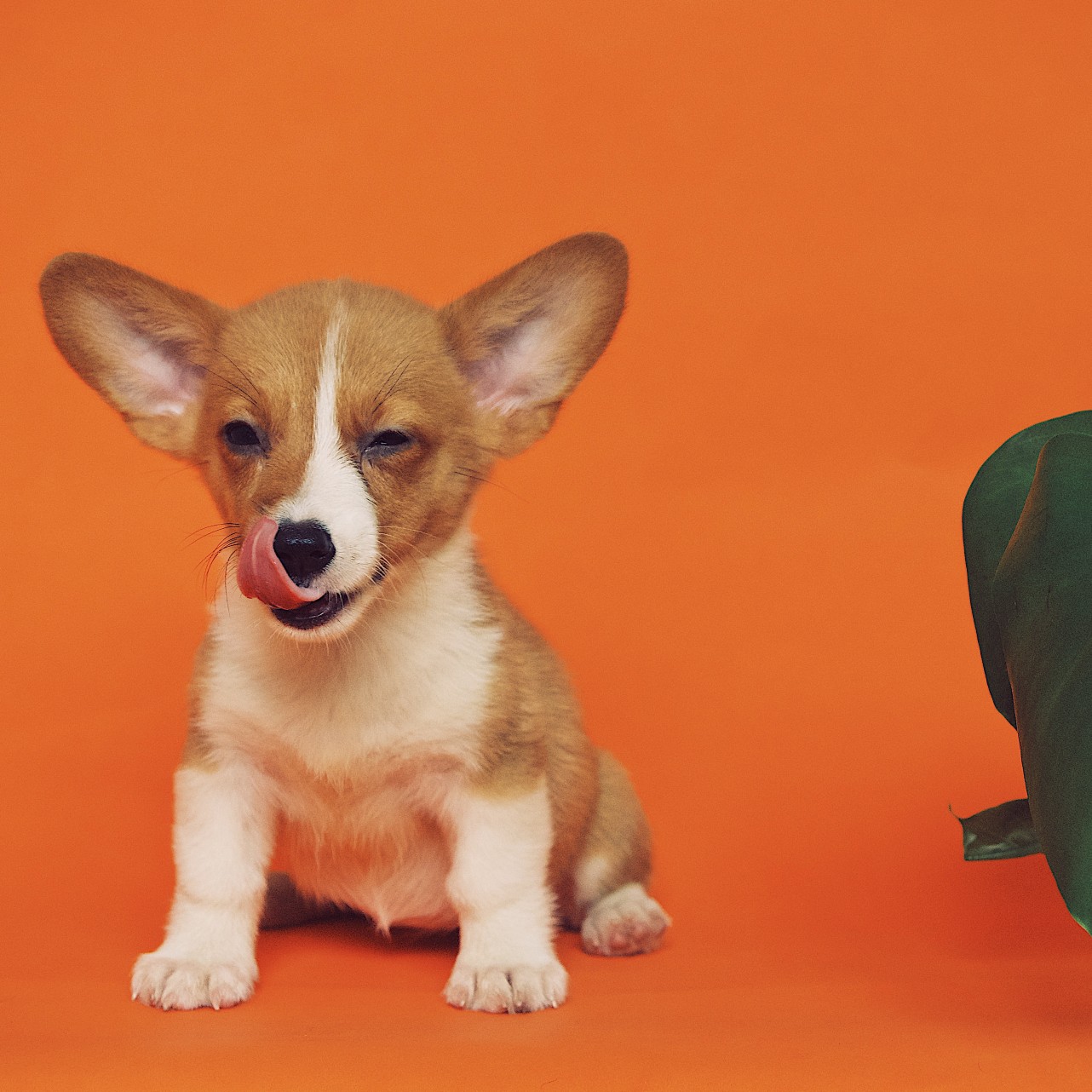} & 
       \includegraphics[width=0.15\linewidth]{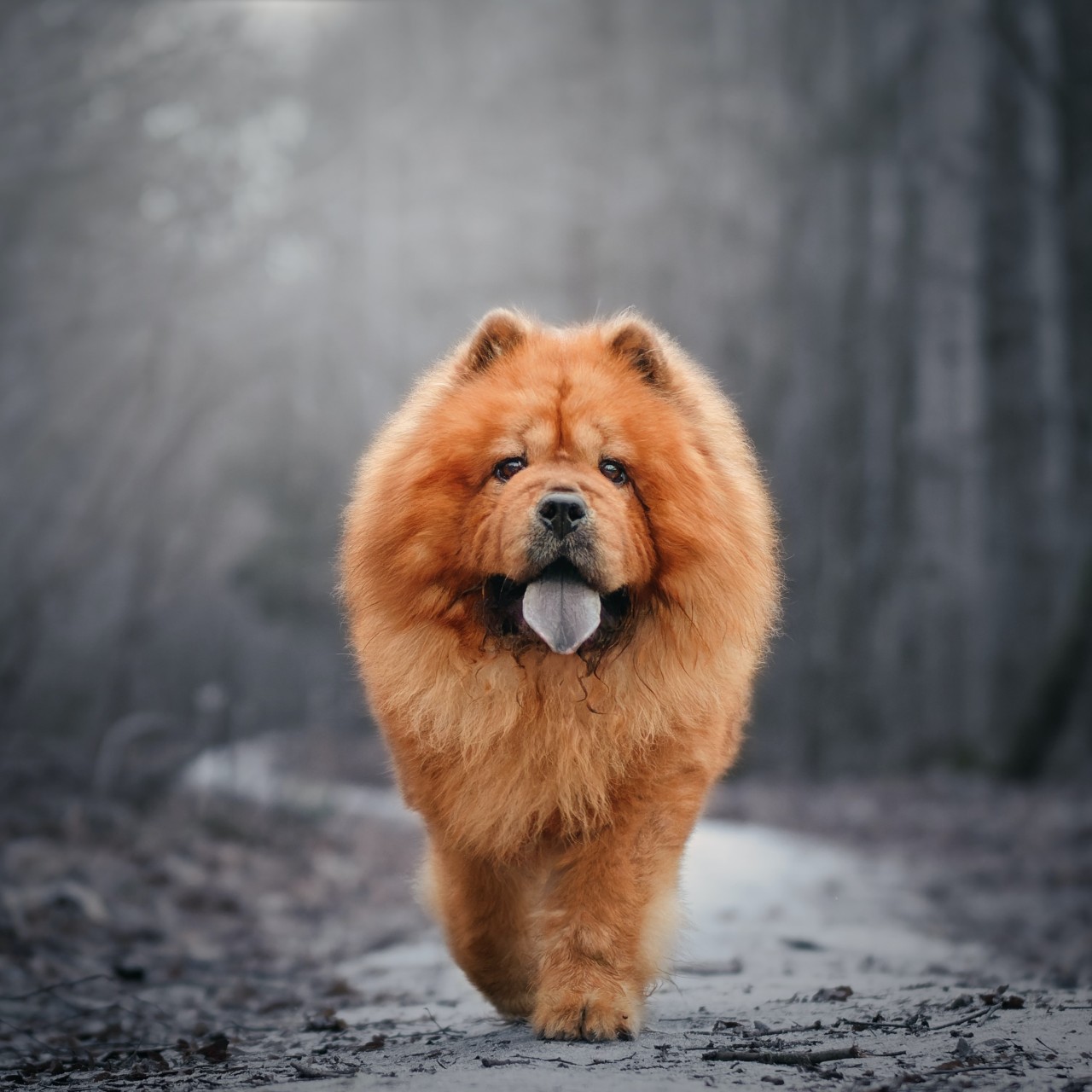} & 
       \includegraphics[width=0.15\linewidth]{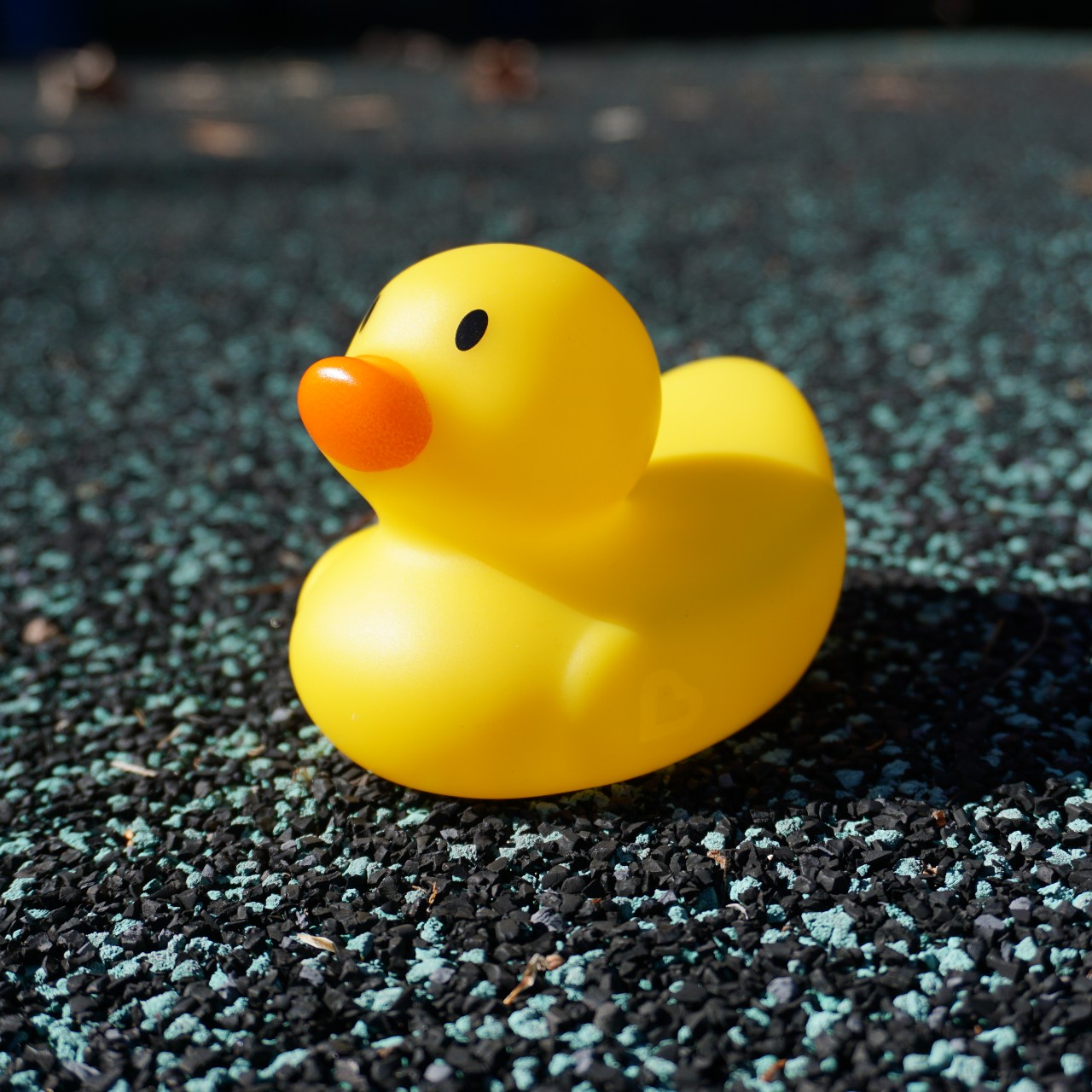}\\
\end{tabular}
\caption{Content images used for the object-style composition evaluation. }
\label{fig:content_images}
\end{figure}

\begin{figure}[ht]
    \centering
    \begin{tabular}{ccccc}
    \includegraphics[width=0.15\linewidth, height=0.15\linewidth]{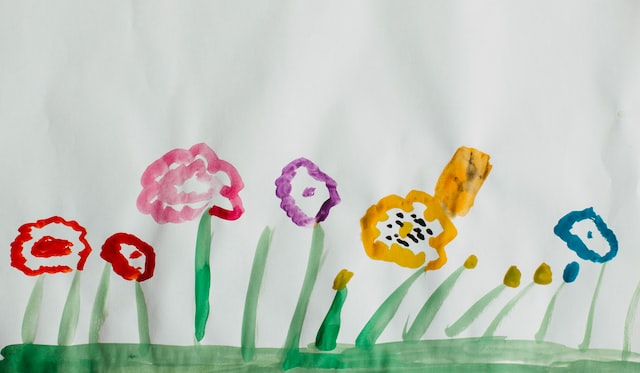} & \includegraphics[width=0.15\linewidth, height=0.15\linewidth]{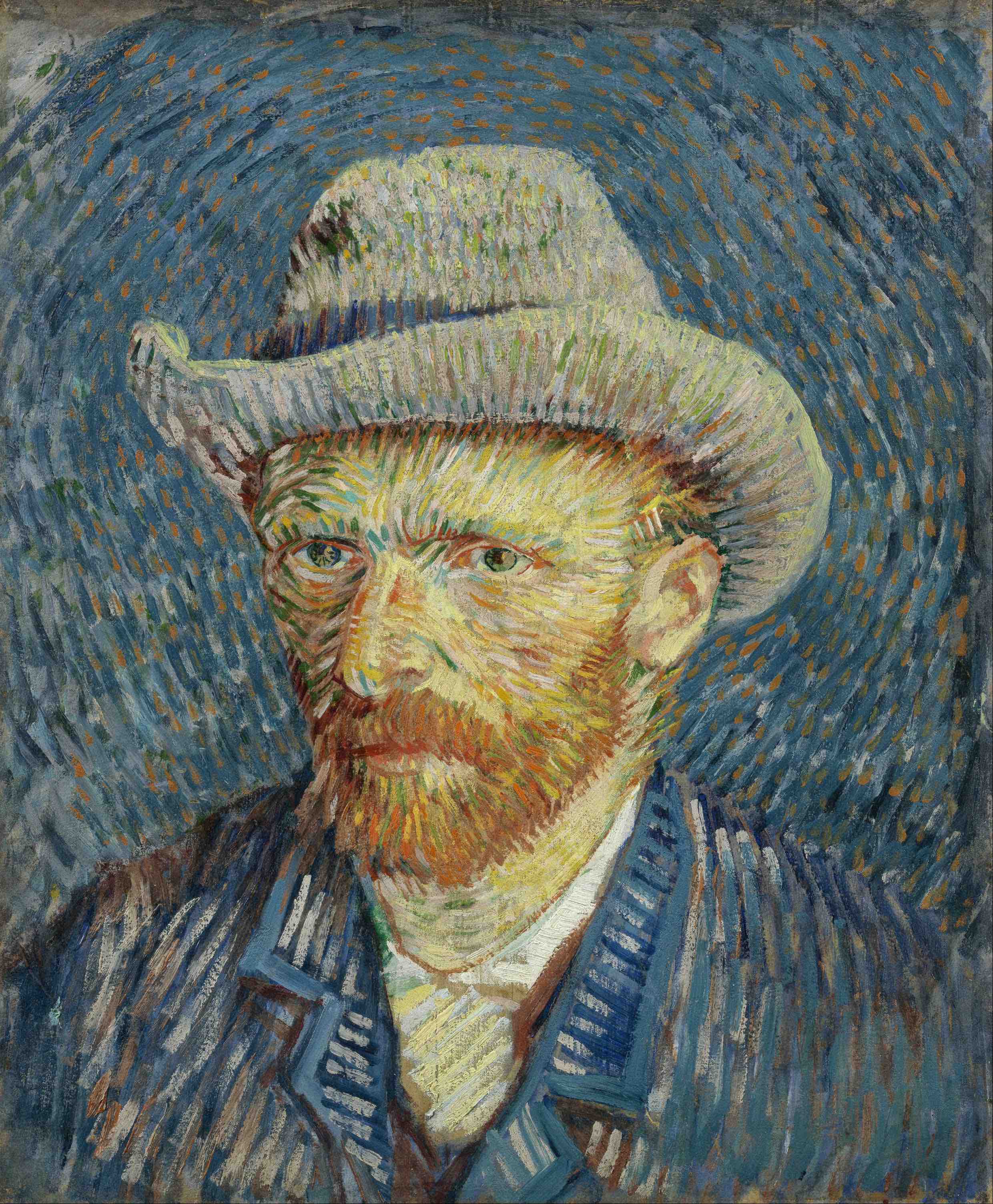} &  \includegraphics[width=0.15\linewidth, height=0.15\linewidth]{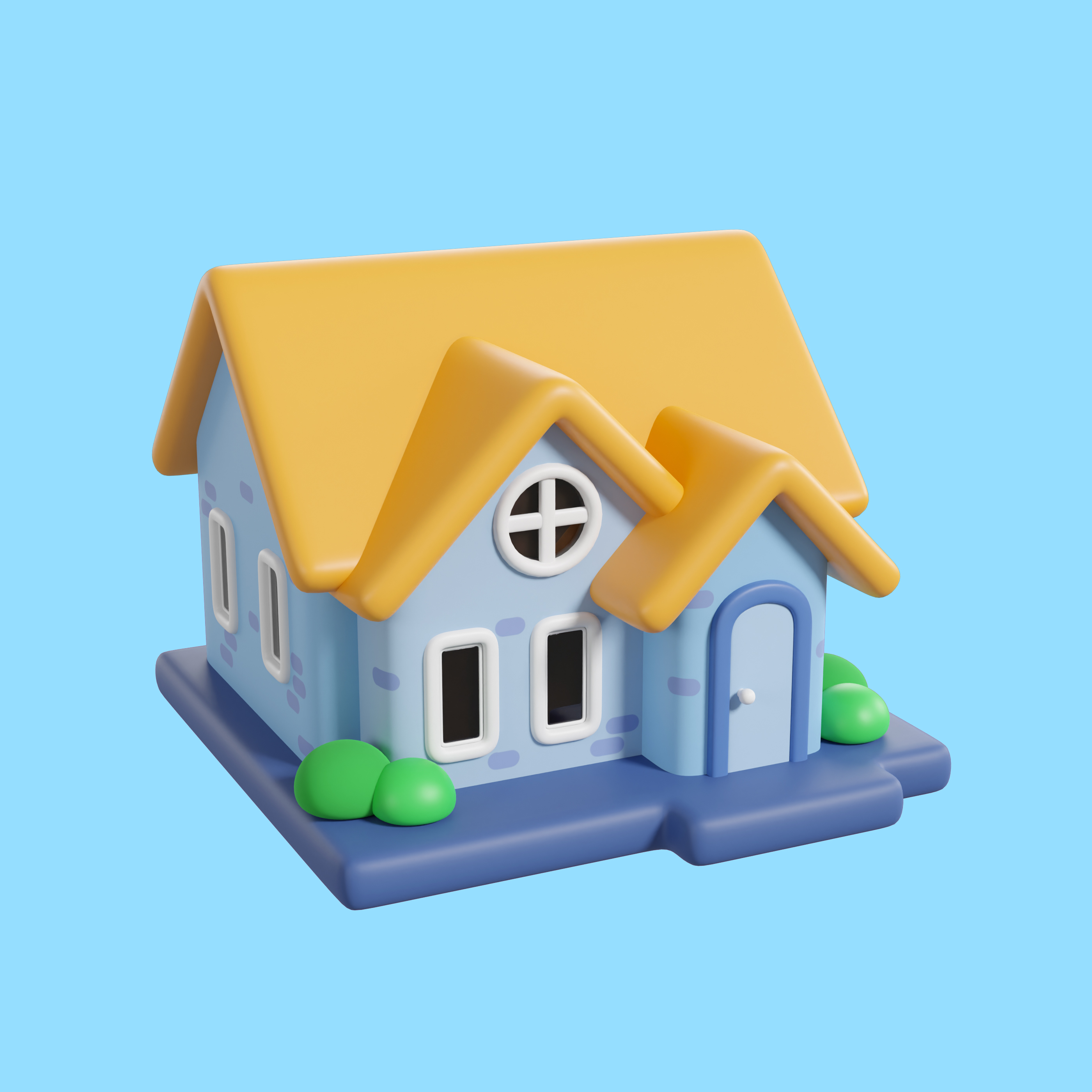} &
     \includegraphics[width=0.15\linewidth, height=0.15\linewidth]{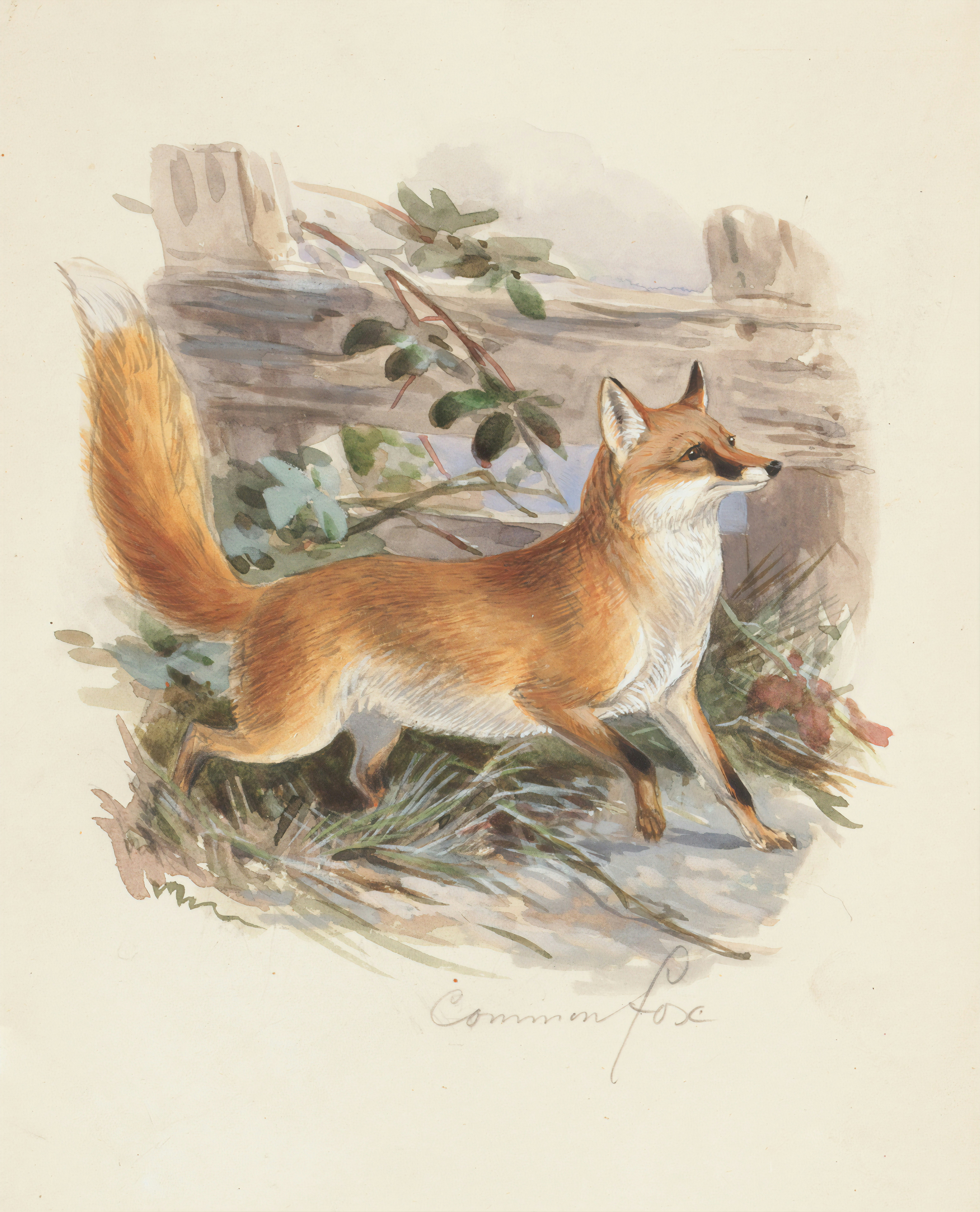} &
      \includegraphics[width=0.15\linewidth, height=0.15\linewidth]{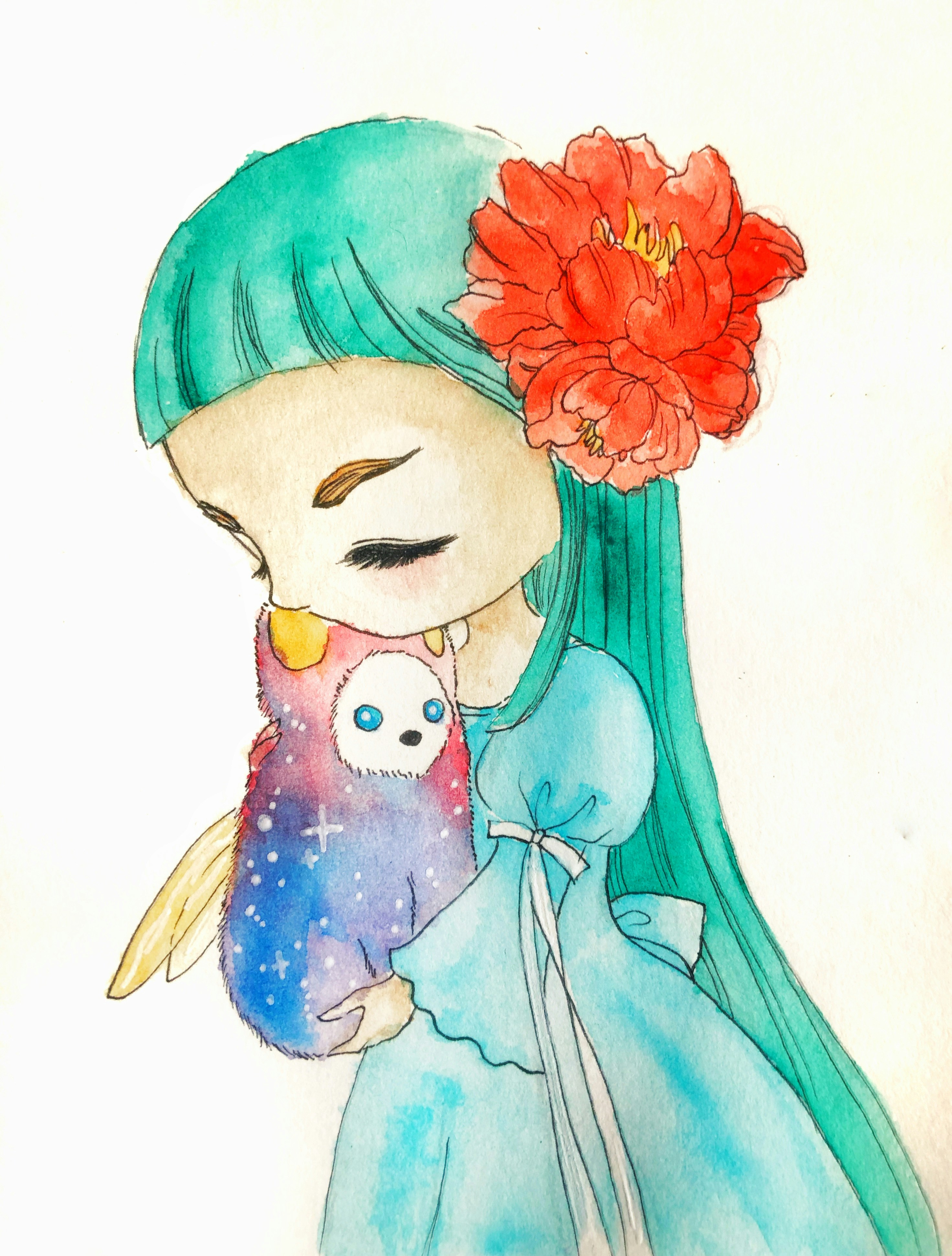} \\
       \includegraphics[width=0.15\linewidth, height=0.15\linewidth]{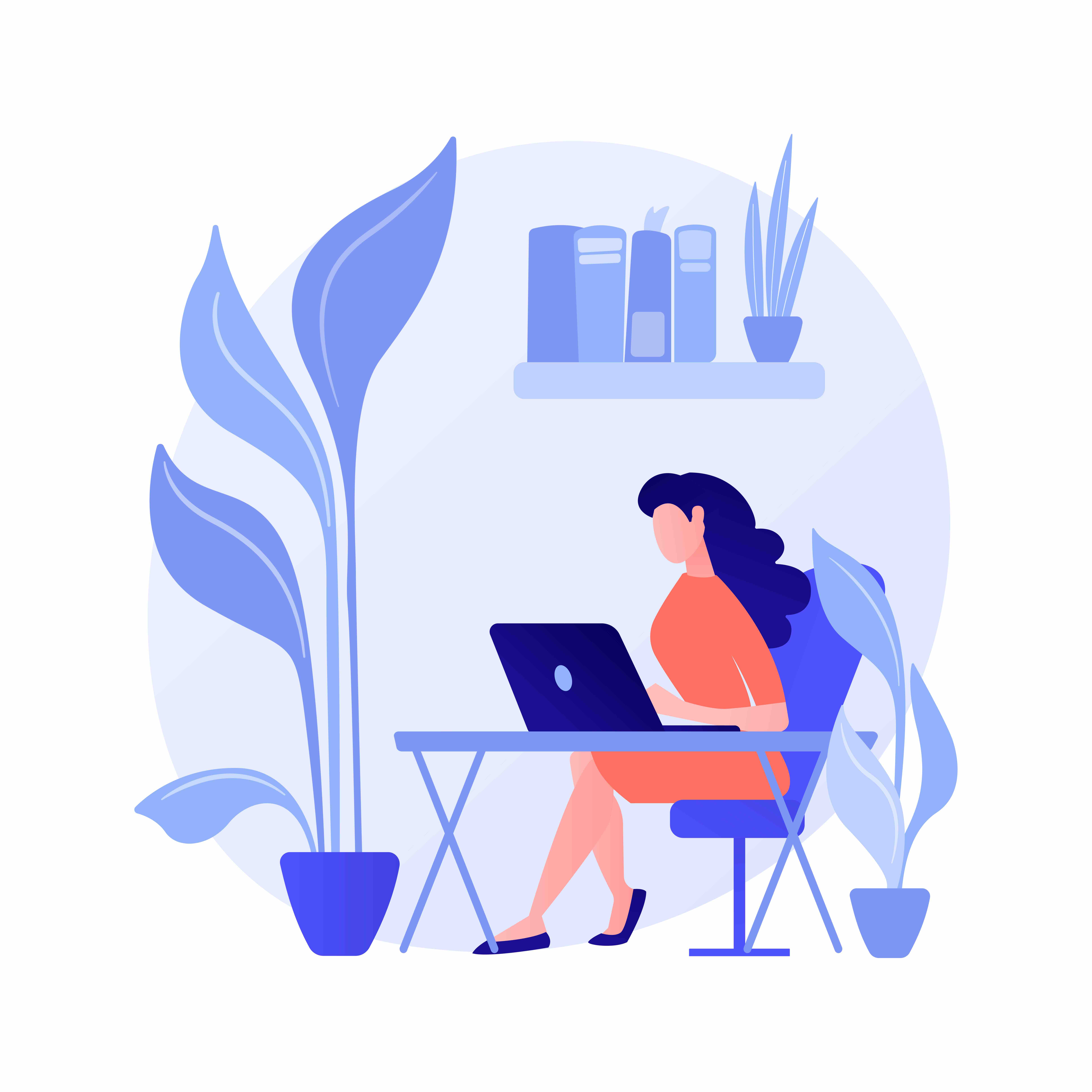} & \includegraphics[width=0.15\linewidth, height=0.15\linewidth]{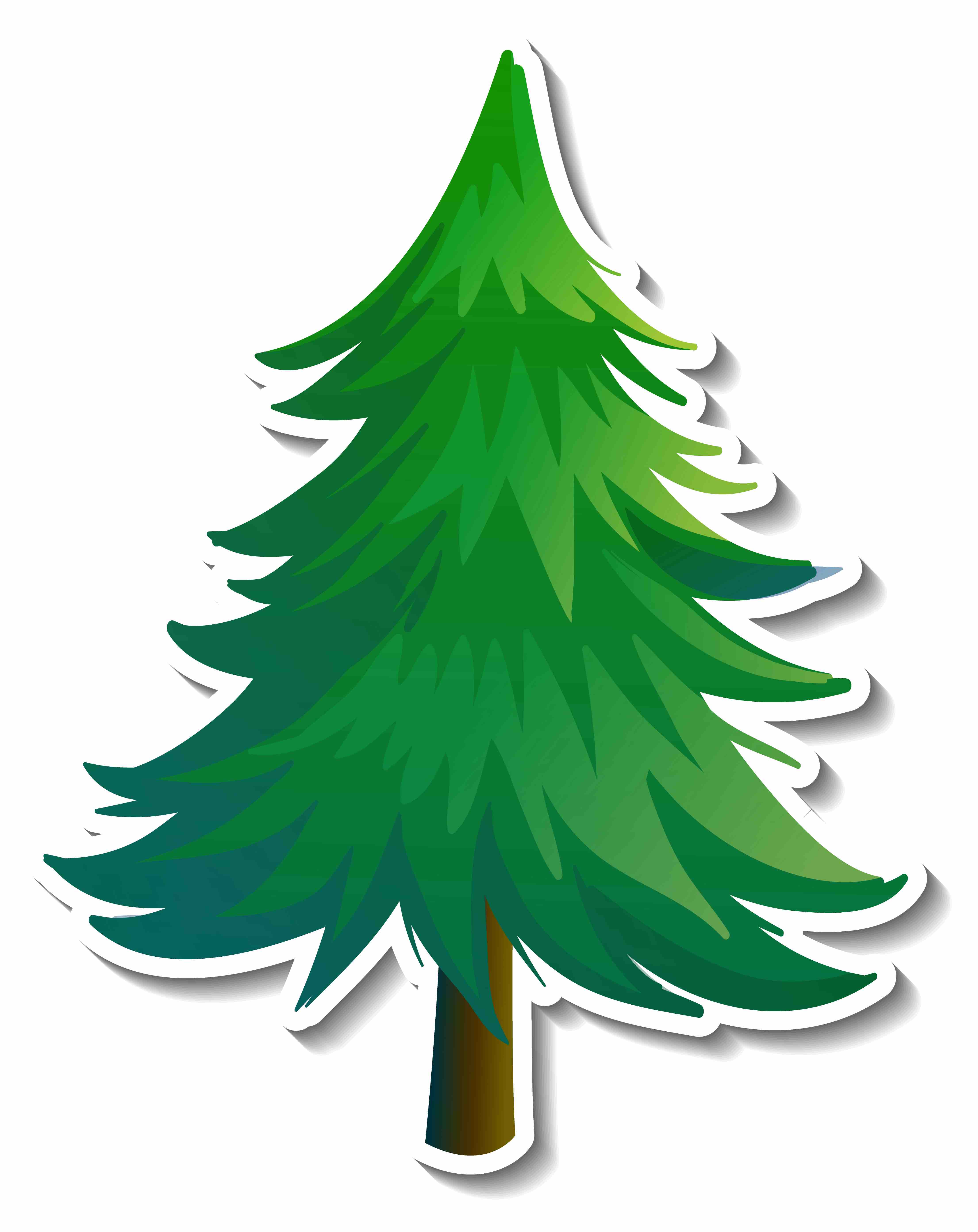} & 
      \includegraphics[width=0.15\linewidth, height=0.15\linewidth]{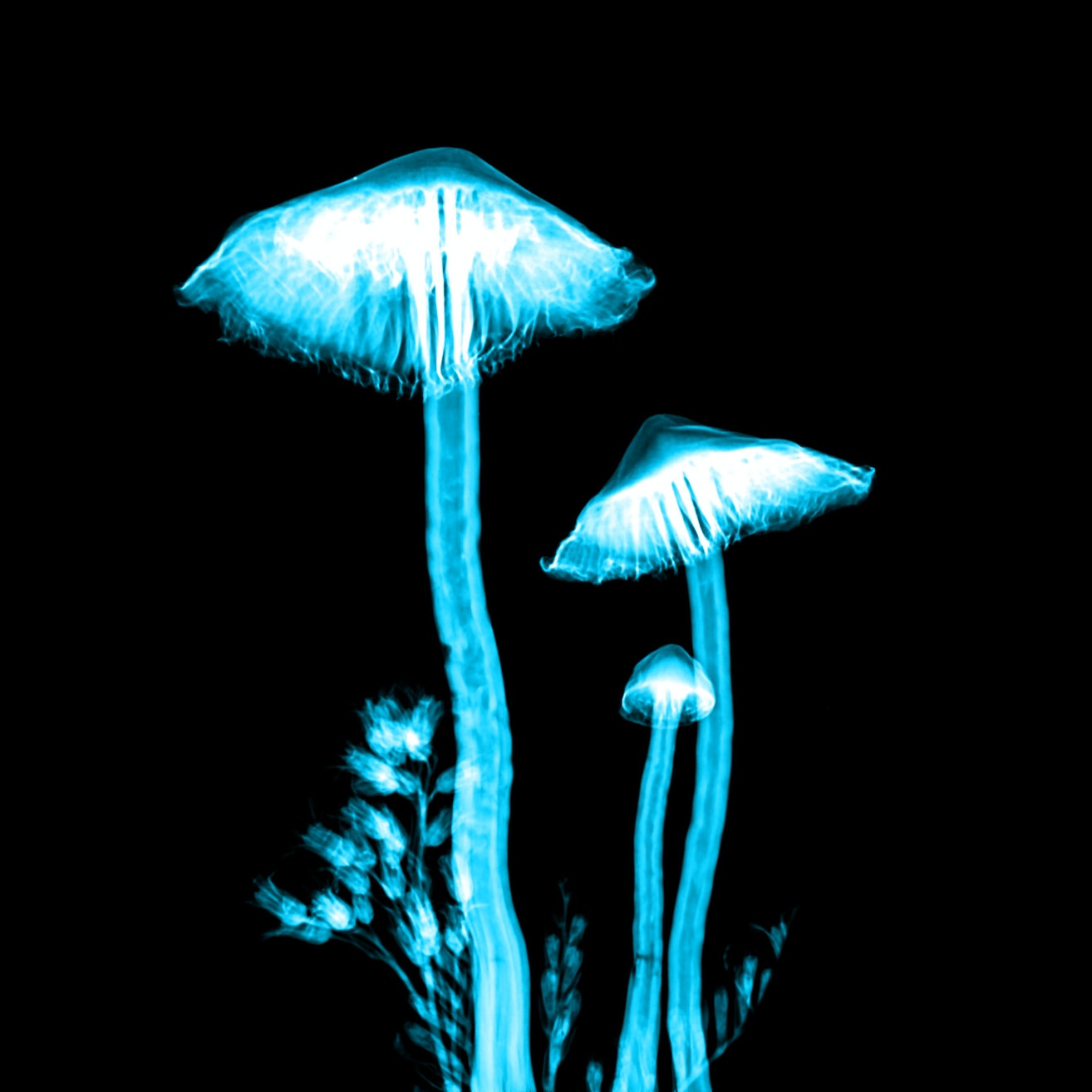} & 
       \includegraphics[width=0.15\linewidth, height=0.15\linewidth]{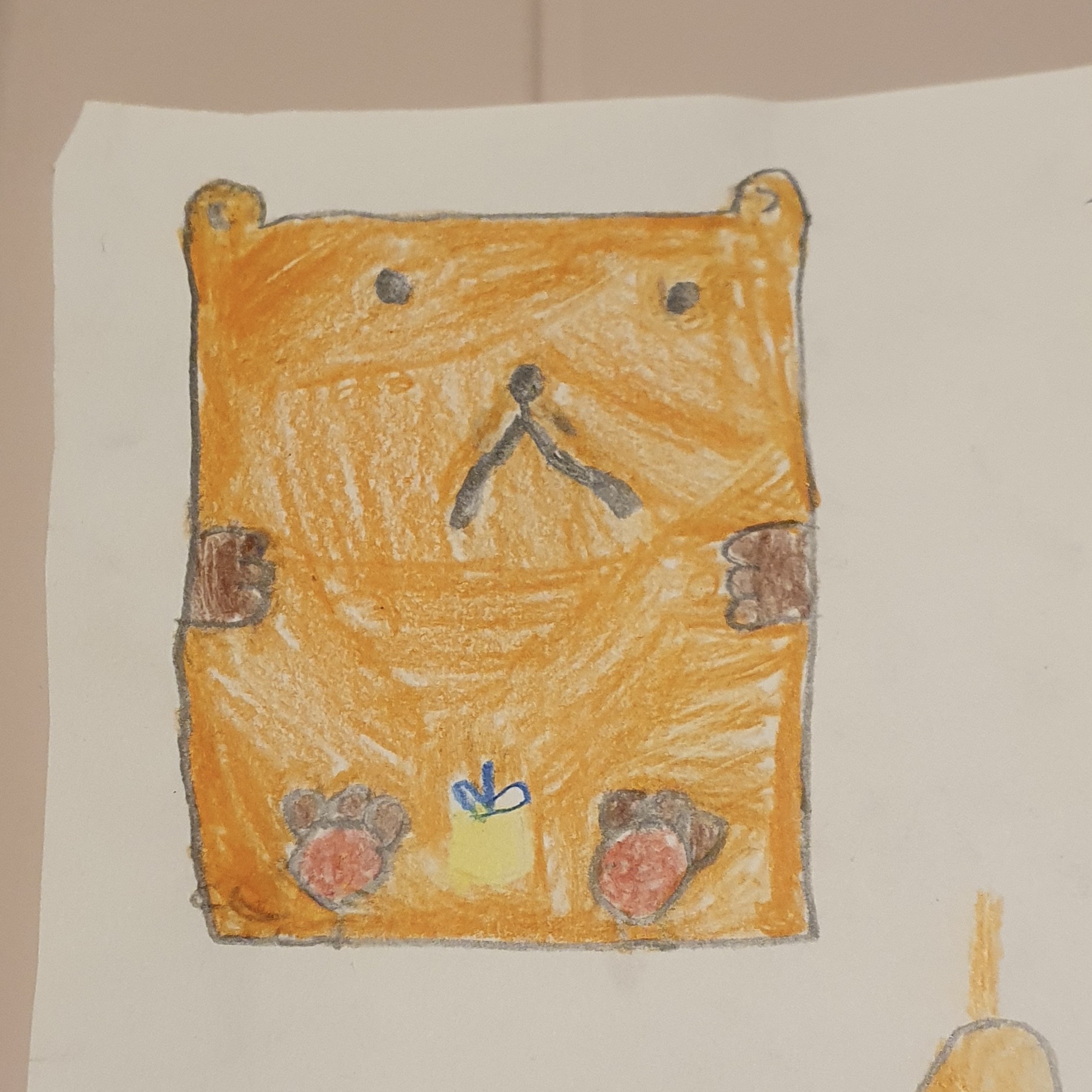} & 
       \includegraphics[width=0.15\linewidth, height=0.15\linewidth]{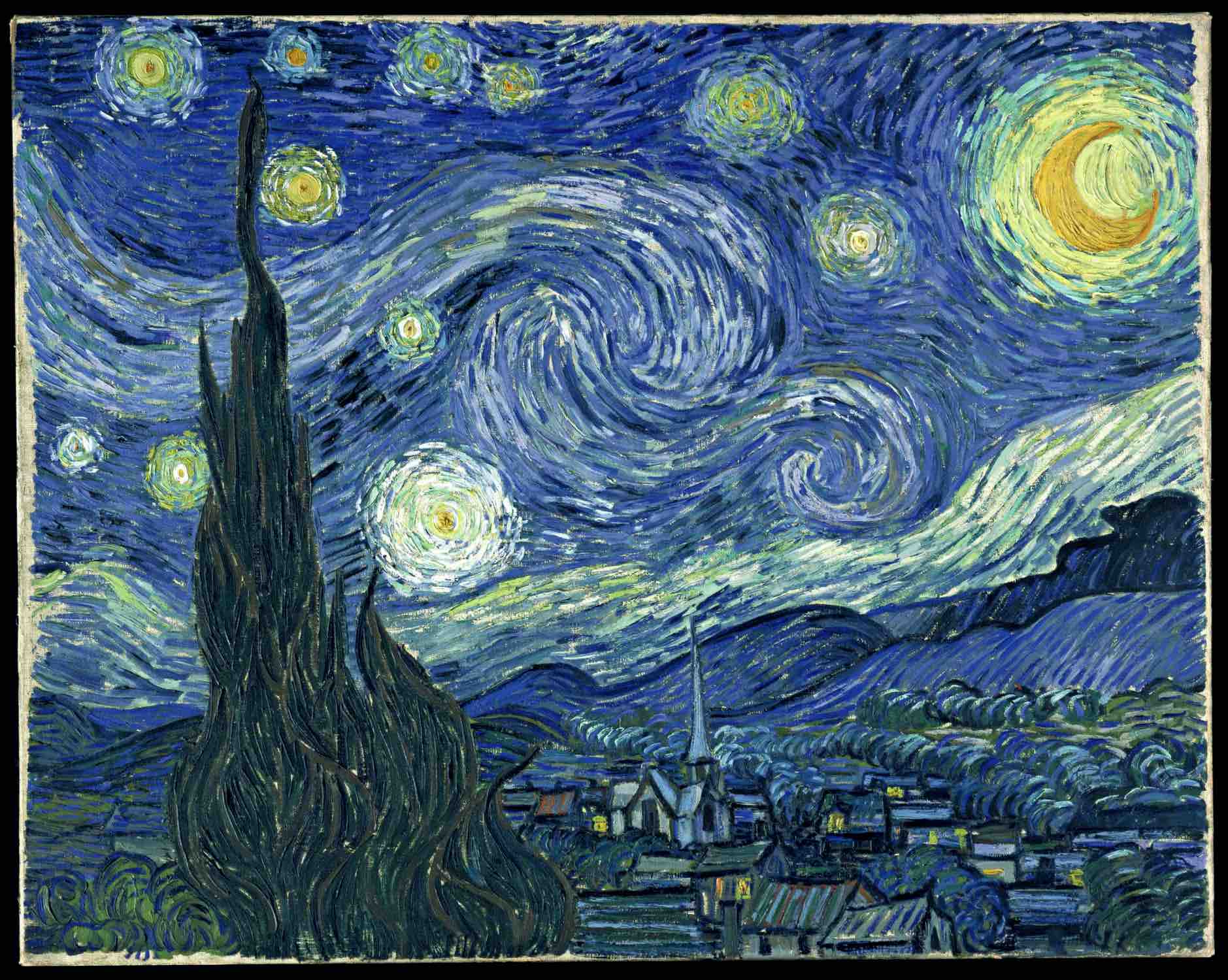}\\
\end{tabular}
\caption{Style images used for the object-style composition evaluation.}
\label{fig:style_images}
\end{figure}

\endgroup

\section{Additional Implementation Details}
\subsection{Training StyleZero and ObjectZero}
We implemented our training pipeline for both StyleZero and ObjectZero using the IP-Adapter~\cite{ye2023ip-adapter} repository\footnote{https://github.com/tencent-ailab/IP-Adapter/tree/main}.  We train both of our adapters for $90$K iterations on four Nvidia A100 GPUs with the batch size of four per each GPU. We train StyleZero using image-text pairs from the ContraStyles~\cite{somepalli2024measuring} dataset and ObjectZero on image-text pairs from MS-COCO~\cite{Lin2014MicrosoftCC}. We use the Adam optimizer with the learning rate of $0.0002$ and weight decay of $0.01$. For both adapters, we set $\gamma$ in the loss as $0.3$.
\subsection{W\"{u}rstchen}
To implement our method (and RB-Modulation) on W\"{u}rstchen architecture, we build on the official codebase\footnote{https://github.com/google/RB-Modulation} provided by RB-Modulation~\cite{rout2024rb} authors. For all experiments, we set $M$ (optimization steps) to $5$. We use a single Nvidia Tesla A100 GPU with batch-size=1. Apart from $M$, we keep the default hyperparameters for RB-modulation intact. To implement SubZero, we set $\gamma_{nc}$ to $1$. For Face-Style (and Action) composition, we set $\gamma_{ns}$ to $0$, and for Object-Style composition experiments, we set $\gamma_{ns}$ to $1$. $\mu_{s,0}$ is set to $0.6$, $\zeta$ is set to $0.4$, and the update is capped once $\mu_{s,t}$ reaches $1$.

\subsection{SDXL-Lightning experiments}
For results on SDXL-Lightning, we implemented all components of SubZero over the official PuLID~\cite{guo2024pulid} repository\footnote{https://github.com/ToTheBeginning/PuLID}, open-sourced by their authors. For face-style composition, we apply various projectors (IP-Adapter\footnote{https://github.com/tencent-ailab/IP-Adapter}, StyleCrafter\footnote{https://github.com/GongyeLiu/StyleCrafter-SDXL} and the proposed StyleZero) for stylization, while keeping the Subject projector as PuLID in all experiments. For object-style composition, we use IP-Adapter, StyleZero and ObjectZero as our style and subject projectors. Unless mentioned otherwise, for weighted aggregation of attention weights, we select the style scales and subject scales which produce the best operating point for all experiments. To report scores with RB-Modulation on SDXL-Lightning, We implement the RB-Modulation stochastic controller in the diffusers pipeline. We set $M$ (optimization steps) to 5. To implement SubZero, we set $\gamma_{nc}$ to $1$. For SubZero Face-Style (and Action) composition, we set $\gamma_{ns}$ to $0$, and for SubZero Object-style composition experiments, we set $\gamma_{ns}$ to $1$. $\mu_{s,0}$ is set to $0.6$, $\zeta$ is set to $0.4$, and the update is capped once $\mu_{s,t}$ reaches $1.5$.

\subsection{Baselines}
\paragraph{InstantID.} To reproduce results using InstantID for subject-style composition, we used an open-source adaptation of their ``Visual Prompting" method\footnote{https://github.com/TheDenk/InstantID-Visual-Prompt/tree/main} on SDXL. We replaced the backbone with SDXL-Lightning and used default settings. We use a single Nvidia Tesla A100 GPU with batch-size $1$.

\paragraph{InstantStyle-Plus.}
We replace the InstantStyle-Plus base model \footnote{https://github.com/instantX-research/InstantStyle-Plus} with SDXL-Lightning while modifying the default settings. For action, we modify the settings for ReNoise to ensure we maintain structural integrity of content and faithfullness to the action specified by prompt while aligning with the style. To ensure action is faithfully generated, we update the number of inversion steps to $40$ and number of renoise iterations per timestep to $4$. In addition, we found that reducing controlnet guidance scale to $0.3$ did not undermine the subject reconstruction. The global and local scales for IP adapter were set at $0.3$ and $0.6$ respectively.

\paragraph{StyleAligned.}
For object-style composition baselines, we replace the base model for StyleAligned \footnote{https://github.com/google/style-aligned/} with SDXL-Lightning while modifying the default settings. Since StyleAligned originally does not input a reference image for style and instead generates the style from a reference prompt, we modify the pipeline to input DDIM inverted latents to the model. The model is conditioned on controlnet. We set the controlnet conditioning scale at $0.9$ and guidance scale at $7.5$. 
We generate images across a single image per prompt for an object-style pair. 

\section{Quantitative Results}
\subsection{Performance of standalone StyleZero and ObjectZero projectors}

\begin{figure}[h]
    \centering
    \includegraphics[width=0.4\linewidth]{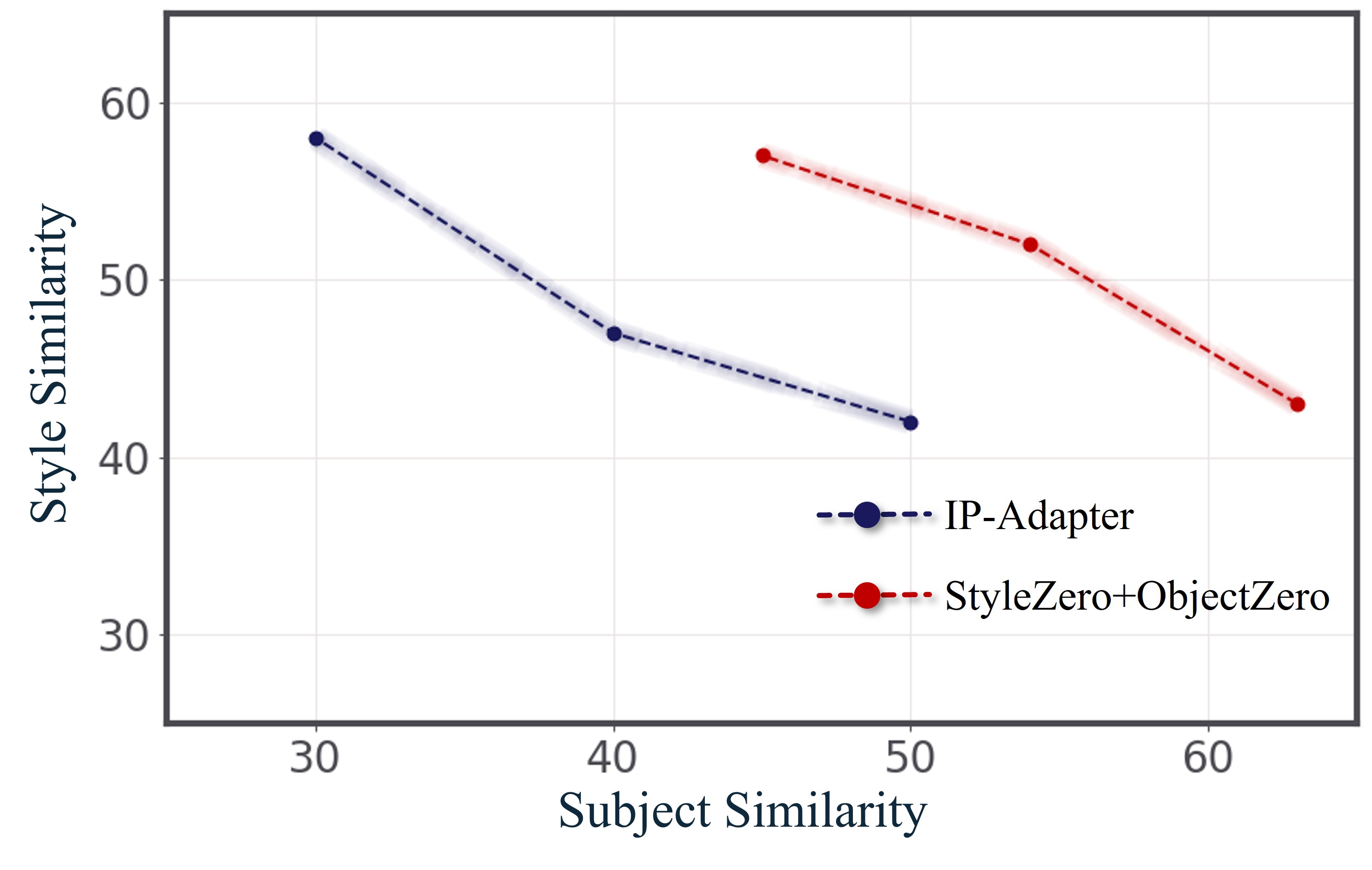}
    \caption{
    % \textbf{Stylized Faces with SubZero}. Stylized Faces with SubZero.
Varying the style and content scaling to generate a trade-off curve between Object and Style similarity for standalone ObjectZero and StyleZero adapters on SDXL.
    }%, and suffer from mode collapse. However, our proposed FouRA produces more diverse images.}
    \label{fig:cross_scale_object}

\end{figure}

Figure~\ref{fig:cross_scale_object} shows the individual gain from our disentangled StyleZero and ObjectZero projector pair over IP-Adapter. We perform this experiment on the object stylization task. However, unlike Table~\ref{tab:obj_stylization_main}, the results are on an SDXL baseline. We compare using our projectors against IP-Adapter. We vary style and subject scaling to generate a trade-off curve between subject and style similarity, on the object-style composition task.
As observed, using the StyleZero and ObjectZero pair provides a significantly better operating point on the Object similarity and Style similarity curve, compared to IP-Adapters. This is due to the fact that our adapters are less prone to Subject leakage.

% \begin{table}[ht]
%     %\addtolength{\tabcolsep}{-1.5pt}
%     \centering
%     \vspace{-0.5 em}

%     \fontsize{7.0pt}{5.75pt}\selectfont
%     \begin{tabular}{c|c|cc}
%     \toprule
%     Style Projector & Object Projector & DINO Sim & Style Sim \\ [0.5ex] 
%     \hline
%     % \midrule
%     % IP-Adapter  & - & 0.156 &  0.756\\
%     % \rowcolor{Gray} StyleZero & - & 0.181 & 0.644 \\
%     % \midrule
%     %  - & IP-Adapter~\cite{ye2023ip-adapter}   & 0.546 & 0.161 \\
%     % % Content & IP-Adapter   & 0.546 & - \\
%     % \rowcolor{Gray} - & ObjectZero (Ours)   & \textbf{0.610} &  \textbf{0.167} \\
%     \midrule
%      IP-Adapter~\cite{ye2023ip-adapter} & IP-Adapter~\cite{ye2023ip-adapter}   & 0.351 & 0.438 \\
%     % Content & IP-Adapter   & 0.546 & - \\
%     \rowcolor{Gray} StyleZero & ObjectZero   & \textbf{0.539} & \textbf{0.441} \\
    
%     % \rowcolor{Gray} Content & ObjectZero   & 0.610 &  - \\
%     \bottomrule
% \end{tabular}
% % \fontsize{8.00pt}{8.25pt}\selectfont

% \caption{\textbf{Our trained standalone object/style adapters}}
% \vspace{-1em}
% \label{tab:standalone}
% \end{table}

\subsection{Varying style and content scaling on Face-Style composition}
\begin{figure}[h]
    \centering
    \includegraphics[width=0.4\linewidth]{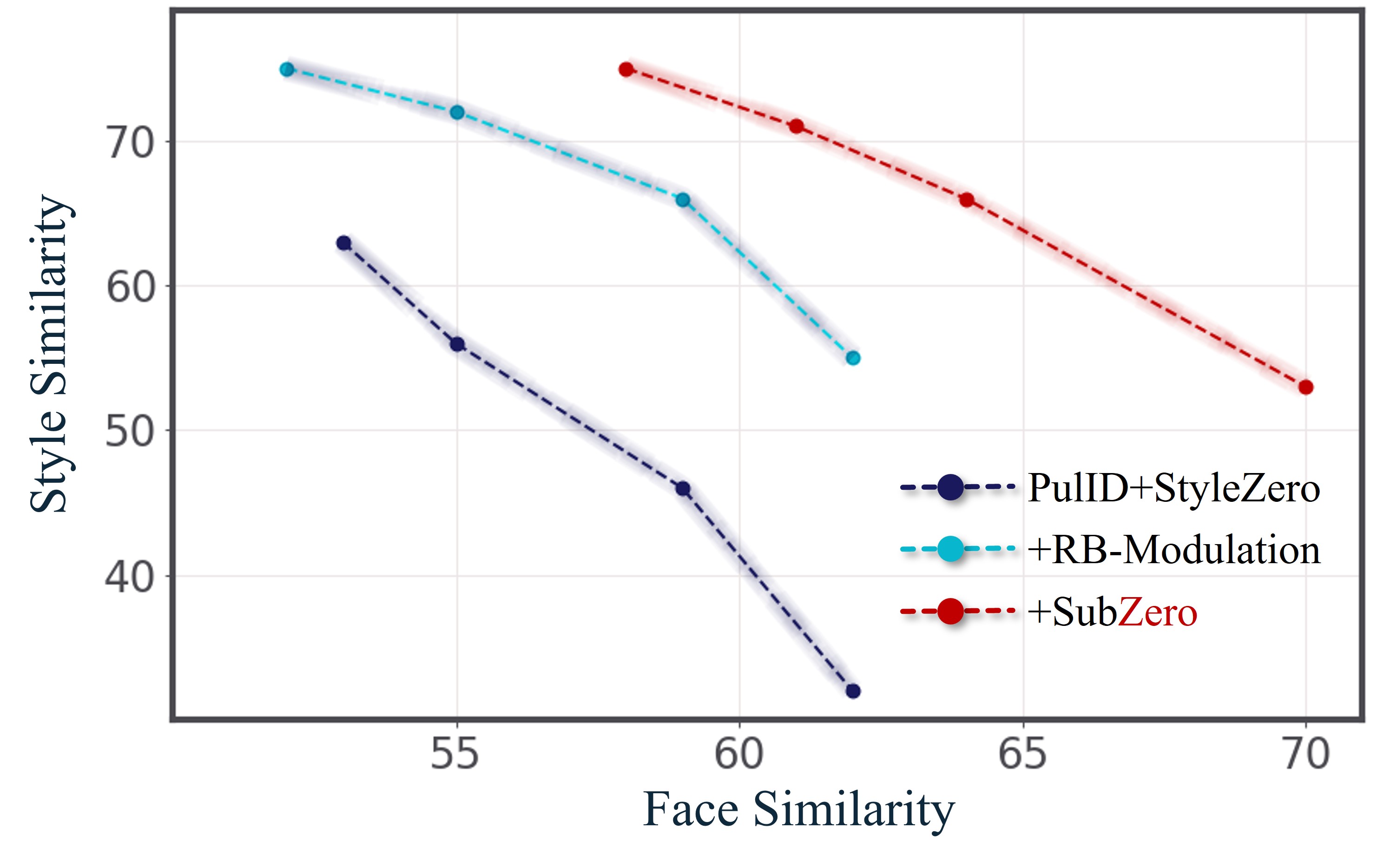}
    \caption{
    % \textbf{Stylized Faces with SubZero}. Stylized Faces with SubZero.
Varying the style and content scaling to generate a trade-off curve between Face and Style similarity.
    }%, and suffer from mode collapse. However, our proposed FouRA produces more diverse images.}
    \label{fig:cross_scale}

\end{figure}

In Figure~\ref{fig:cross_scale}, we vary style and content scaling to generate a trade-off curve between face and style similarity, on the face style composition task. All results are without style helper prompts on SDXL-Lightning, as an extension of the ones shown in Table~\ref{tab:face_stylization_main}. We compare our StyleZero projector added to PuLID, RB-Modulation (with both these projectors) and our proposed SubZero approach. As observed, SubZero observe a consistent improvement over RB-Modulation and naiive merging of base projectors over a distribution of scales.

\subsection{Subject leakage measurement}
\begin{figure}[h]
    \centering
    \includegraphics[width=0.8\linewidth]{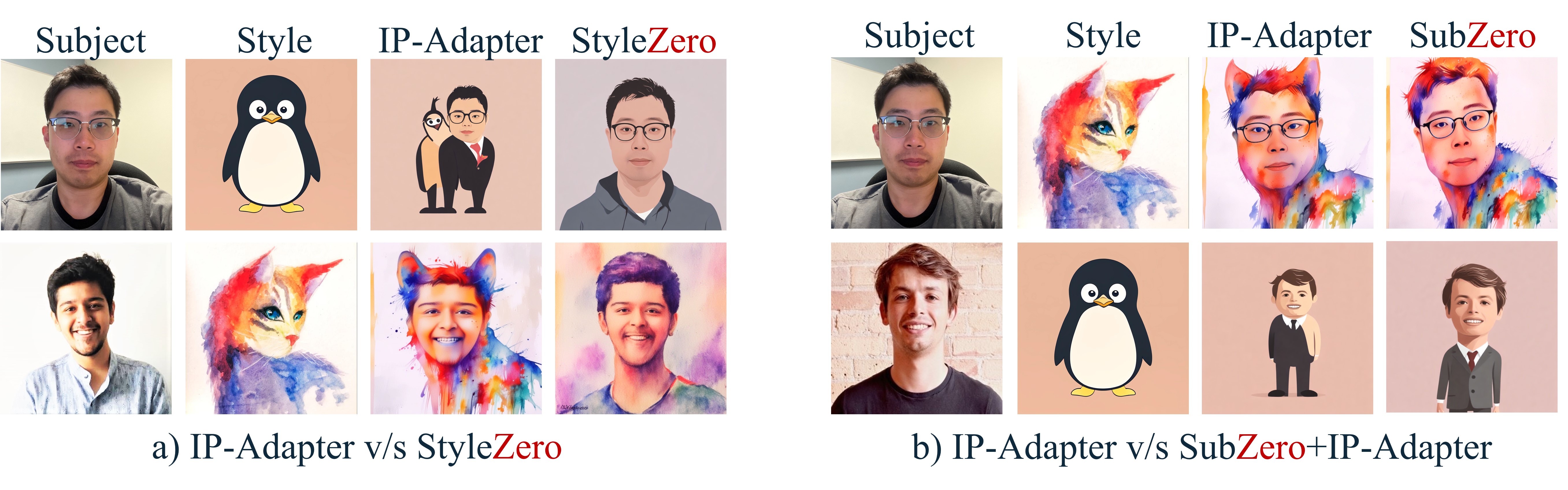}
    \caption{
    % \textbf{Stylized Faces with SubZero}. Stylized Faces with SubZero.
Visualizing subject leakage for various schemes
    }%, and suffer from mode collapse. However, our proposed FouRA produces more diverse images.}
    \label{fig:sub_leakage}

\end{figure}

To effectively quantify and measure subject leakage, we curate a dataset of 10 style images which are likely susceptible to leakage. This dataset is described in Section~\ref{subsec:data_app}. To measure leakage, we measure a normalized CLIP similarity between generated images and the leakage text prompts. We show quantitative results in Table~\ref{table:subj_leakage}, and qualitative results in Figure~\ref{fig:sub_leakage}. As shown from results, the StyleZero projector significantly reduces leakage while keeping the Style Similarity consistent. Additionally, SubZero the inference algorithm including OTA and Disentangled Latent Optimization further improves subject and style similarity, while reducing leakage. This is also evident in Figure~\ref{fig:sub_leakage}, as subject leakage artifacts, which include cat ears, dog ears and subject shape are fixed by either the StyleZero projector and SubZero inference.

\label{subsec:leakage}

\begin{table}[h]
    %\addtolength{\tabcolsep}{-1.5pt}

    \centering
    \fontsize{7.0pt}{5.75pt}\selectfont
    \begin{tabular}{c|cc|ccc}
    \toprule
    Style projector & Disentangled Control & Ortho. Temporal Aggregation & Face Sim.($\uparrow$) & Style Sim.($\uparrow$) & Subject Leakage($\downarrow$)\\
    \hline
    \midrule
        &  & & 56.2 & 59.1 & 54.6 \\
       \rowcolor{Gray} &  & \checkmark & 58.3  & 58.7 & 41.5 \\
       %\midrule
       %RB-Modulation & & & 54.1 & \textbf{76.5} & 58.3 \\
       \rowcolor{Gray} \multirow{-3}{*}{IP-Adapter} & \checkmark & \checkmark & \textbf{64.8} & \textbf{70.1}  & \textbf{33.4}  \\
    \midrule
       \rowcolor{Gray} &  & & 57.2 & 60.8 & 55.4   \\
       \rowcolor{Gray}  &  & \checkmark & 60.3 & 59.0 & 37.6 \\
       %\midrule
       %RB-Modulation & & & 54.1 & \textbf{76.5} & 58.3 \\
       \rowcolor{Gray} \multirow{-3}{*}{StyleZero} & \checkmark & \checkmark & \textbf{66.4} & \textbf{69.3} & \textbf{28.6} \\

      \hline
    \bottomrule
\end{tabular}
% \fontsize{8.00pt}{8.25pt}\selectfont
\vspace{-0.7 em}
\caption{\textbf{Measuring Subject Leakage:} We report results on SDXL-Lightning with IP-Adapter and PulID. All numbers are without style helper prompts.}\label{table:subj_leakage}
\vspace{-1.0 em}
\end{table}

\subsection{Runtime Analysis}
Table~\ref{table:latency} lists the overall runtime to generate face-style composed images with SDXL-Lightning baseline. All numbers are using style helper prompts. The measurements are on a single Nvidia A100 GPU. As observed, the Orthogonal Temporal Aggregation and Disentangled Stochastic Optimal Control algorithms trade-off performance in terms of Face and Style similarity, with latency. For a gradient-free inference suitable for mobile devices, our StyleZero adapter with Orthogonal Temporal Aggregation of attention features achieves the most promising operating point. This method can also successfully reduce subject leakage, as shown in Table~\ref{table:subj_leakage}.

\begin{table}[h]
    %\addtolength{\tabcolsep}{-1.5pt}

    \centering
    \fontsize{7.0pt}{5.75pt}\selectfont
    \begin{tabular}{ccc|ccc|c}
    \toprule
    StyleZero & Ortho Temporal Aggregation & Dis. Control & Face Sim. & Style Sim. & Average & Runtime (sec)\\
    \hline
    \midrule
     \rowcolor{Gray} \checkmark & & & 59.5 & 58.4 & 59.0 & 0.7 \\
      \rowcolor{Gray} \checkmark & \checkmark &  & 60.1 & 61.9 & 61.0 & 0.9 \\
      \rowcolor{Gray} \checkmark & \checkmark & \checkmark & \textbf{66.5} & \textbf{72.4} & \textbf{69.5} & 2.0 \\
    \bottomrule
\end{tabular}
% \fontsize{8.00pt}{8.25pt}\selectfont
\caption{\textbf{Runtime Analysis from SubZero components:} We report total runtime with results on SDXL-Lightning with StyleZero. All numbers are with style helper prompts}\label{table:latency}

\end{table}

\subsection{Zero-Order Stochastic Optimal Control}

As discussed in Section~\ref{subsec:zo}, Zero-Order(ZO) methods approximate the gradient by perturbing the weight parameters by a small amount based on some random noise.
As shown in Table~\ref{tab:zo_control}, we perform preliminary experiments by leveraging the ZO-Adam scheme described in MeZO ~\cite{malladi2024finetuninglanguagemodelsjust} and extend it to update the latent in the optimizer. This experiment is on the W\"{u}rstchen architecture, performing Face-Style composition for 4 subjects and 30 styles over a single seed. We report the Face Similarity metric along with cached memory overhead for backpropagation, $\Delta_{bp}$.  For this experiment, we focus on a single constraint, i.e. the subject descriptor constraint $\mathcal{L}_c$ from Equation~\ref{eq:dsoc}. This is due to the fact that gradient-free methods find it harder to converge with additional criterions. The first row provides performance and $\Delta_{bp}$ measurements on base Wurschten model without stochastic control. The second row shows results with gradient descent(as used in our paper), and the third row shows zero-order optimization. As stated in the table, we observe that while ZO optimization is not at par with gradient descent, it shows that it outperforms the base model with no latent optimization - achieving a competitive personalization distance.
Also, the memory savings resulting from ZO are significant. Thus, we suggest the use of ZO techniques for the latent update in scenarios where one can afford to trade training time for a more favorable memory budget. Our experiments with ZO are preliminary, and moving forward we intend to explore this area in much more detail.

\begin{table}[ht]
    %\addtolength{\tabcolsep}{-1.5pt}
    \centering
    \vspace{-0.5 em}

    \fontsize{7.0pt}{5.75pt}\selectfont
    \begin{tabular}{cc|cc}
    \toprule
    Latent Optimization & Zero Order & Face Sim & $\Delta_{bp}$ (GB) \\ [0.5ex] 
    \hline
    \midrule
     &  & 57.7 & 0 \\
    % Content & IP-Adapter   & 0.546 & - \\
     \rowcolor{Gray} \checkmark & & \textbf{65.4} & 5.6 \\
     \rowcolor{Gray} \checkmark & \checkmark & 58.9 & 0 \\
    
    % \rowcolor{Gray} Content & ObjectZero   & 0.610 &  - \\
    \bottomrule
\end{tabular}
% \fontsize{8.00pt}{8.25pt}\selectfont

\caption{\textbf{Zero-Order Stochastic Controller}}
\label{tab:zo_control}
\end{table}

\section{Qualitative Results}
\subsection{Face-Style Composition}

\begin{figure}[h]
    \centering
    \includegraphics[width=1.0\linewidth]{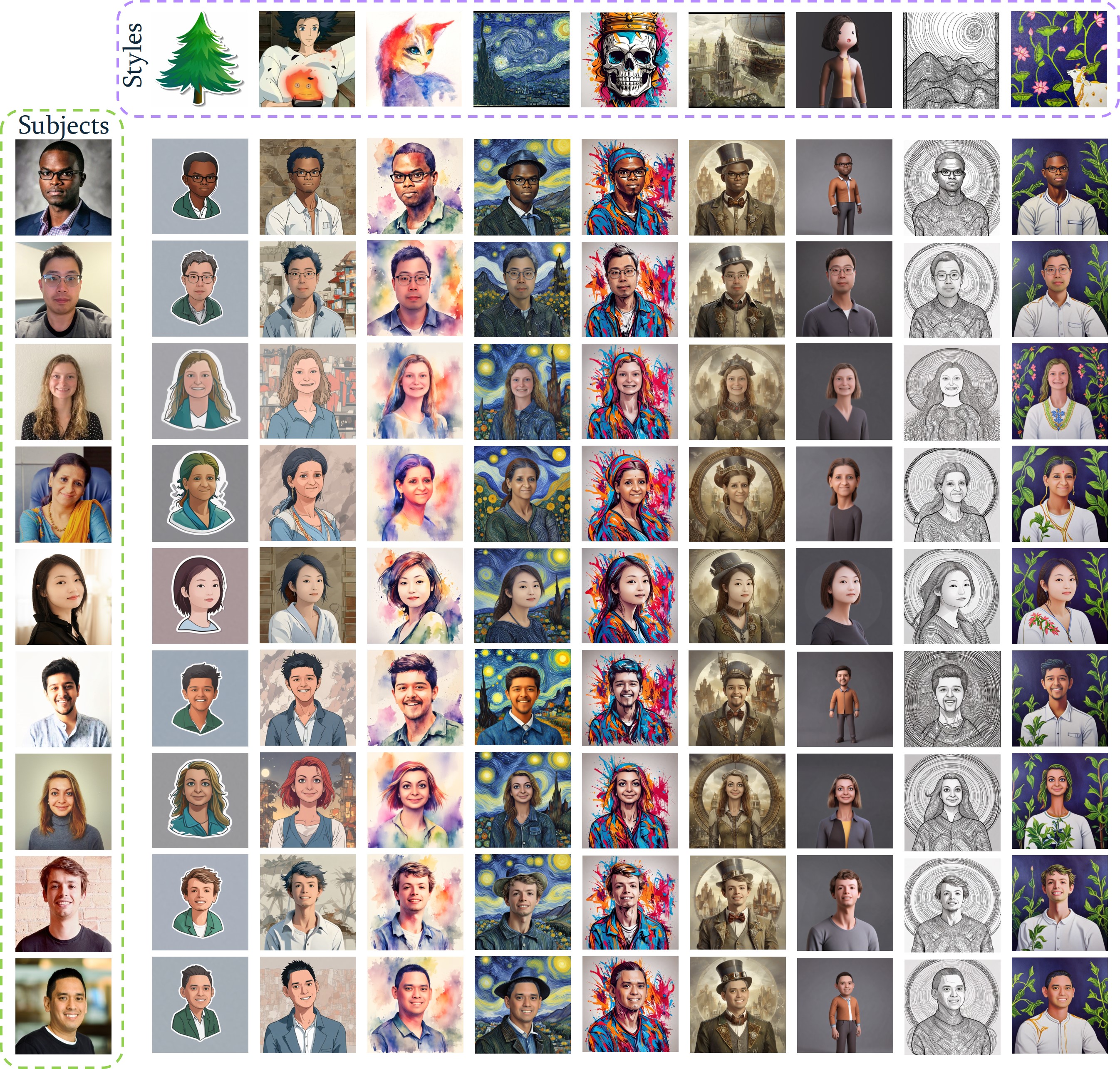}
    \caption{
    % \textbf{Stylized Faces with SubZero}. Stylized Faces with SubZero.
Various stylized face images generated using our proposed SubZero method. These images are using style helper prompts.
SubZero produces high-quality, diverse stylized images while maintaining facial features.
    }%, and suffer from mode collapse. However, our proposed FouRA produces more diverse images.}
    \label{fig:faces_stylized_all}

\end{figure}

\begin{figure}[h]
    \centering
    \includegraphics[width=1.0\linewidth]{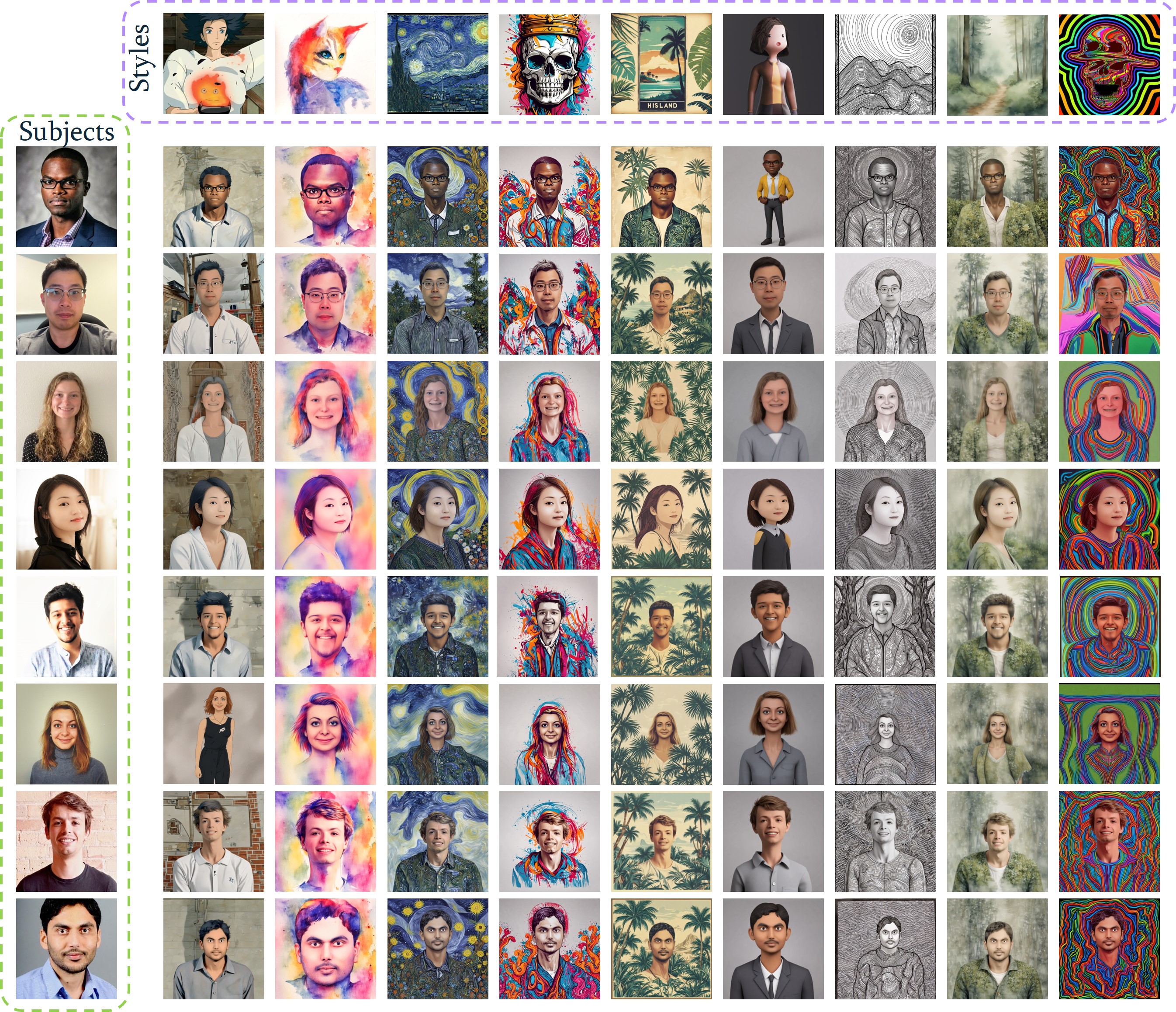}
    \caption{
    % \textbf{Stylized Faces with SubZero}. Stylized Faces with SubZero.
Various stylized face images generated using our proposed SubZero method. These images are without style helper prompts. Even without style descriptors in the prompt, SubZero produces images which remain faithful to the input style while maintaining facial features.
    } \label{fig:faces_stylized_nohelpers}

\end{figure}

Figure~\ref{fig:faces_stylized_all} is an extension to our Fig~\ref{fig:faces_stylized}, and shows SubZero results for 9 faces stylized by 9 styles. As observed, SubZero can stylize a wider distribution of faces across a broad range of styles in a zero-shot setting. These images are generated with style descriptor prompts. Additionally, we show SubZero face-style composition results without style helper prompts in Figure~\ref{fig:faces_stylized_nohelpers}. As observed, our trained StyleZero adapter can effectively adapt to a wide variety of styles, without the need for the style descriptor in prompt. This is an elusive goal in the domain of image stylization, as also discussed by the authors of~\cite{rout2024rb} and ~\cite{shah2025ziplora}.

\clearpage

\subsection{Object-Style Composition}
\begin{figure}[h]
    \centering
    \includegraphics[width=1.0\linewidth]{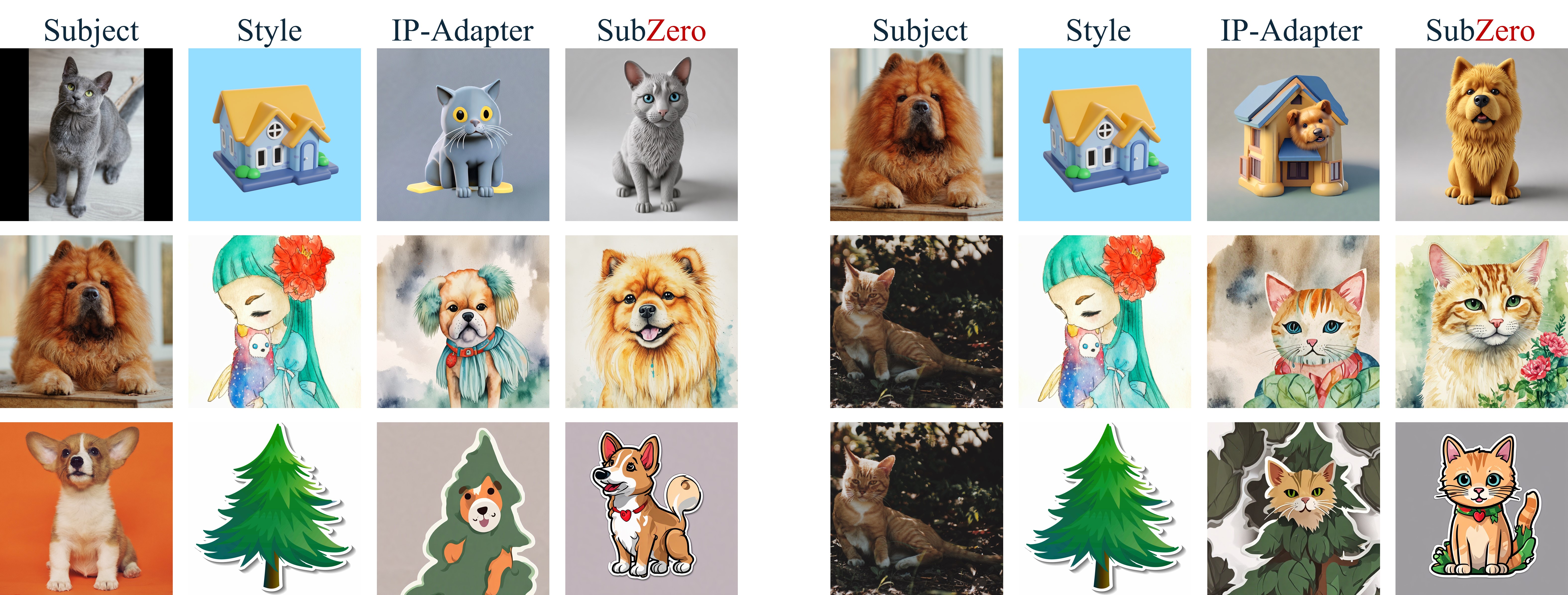}
    \caption{
    % \textbf{Stylized Faces with SubZero}. Stylized Faces with SubZero.
SubZero object-style composition v/s IP-Adapter. All results are using SDXL lightning backbone. As observed, IP-Adapter contains subject leakage artifacts, which are clearly fixed when using SubZero.
    } \label{fig:object_style_app}

\end{figure}

Figure~\ref{fig:object_style_app} is an extension to our Fig~\ref{fig:object_style_action}, and shows SubZero results for object-style composition compared to IP-Adapter. As clearly visible in the image, IP-Adapter contains subject leakage artifacts, which are clearly fixed when using SubZero.

\section{Limitations and Future Work}

While SubZero manages to produce a significant improvement in performance on Subject, Style and Action composition over current SOTA, we observe that there is still a scope for improvement. In certain cases with detailed action prompts, we observe artifacts such as multiple-object generation and distortion. This is also attributed to the fact that SDXL-Lightning is a 4-step diffusion model, and does not enable corrective negative prompting with guidance conditioning. Hence, we aim to improve the robustness of this method by integrating newer baselines which produce lesser failure cases.

Furthermore, our proposed zero-order optimization for latent optimization is a promising step to incorporate zero-order training within the vision community. While our method can run on a mobile device without latent optimization, we plan to build on our ZO results to enable the capabilities of our proposed disentangled stochastic optimal controller for mobile devices which cannot perform back-propagation.

Overall, assessing the performance of SubZero, we believe that our proposed method will lay a foundation for further research in training-free personalization

\end{document}